\documentclass[journal,twoside,web]{ieeecolor}
\usepackage{tmi}
\usepackage{cite}
\usepackage{amsmath,amssymb,amsfonts}
\usepackage{algorithmicx}
\usepackage{url}
\usepackage{graphicx}
\usepackage{textcomp}
\usepackage{xcolor}

\usepackage{hyperref}
\usepackage{booktabs}
\usepackage{multirow}
\usepackage{multicol}
\usepackage{arydshln}
\usepackage{kotex}
\usepackage[ruled, vlined, commentsnumbered, linesnumbered]{algorithm2e}

\usepackage[flushleft]{threeparttable}

\newcommand{\eg}{{\em e.g.}}           % e.g.
\newcommand{\ie}{{\em i.e.}}           % i.e.
\newcommand{\etc}{{\em etc}}         % etc.

\newcommand{\bX}{{\mathbf{X}}}         % etc.

\newcommand{\rev} {\color{black}}
\newcommand{\revv} {\color{black}}

\def\BibTeX{{\rm B\kern-.05em{\sc i\kern-.025em b}\kern-.08em
    T\kern-.1667em\lower.7ex\hbox{E}\kern-.125emX}}
\begin{document}
% \title{Inter-regional High-level Relation Learning and Personalized Regions Selection from Functional Connectivity via Self-supervision}
% \title{LRP-RS: A Novel \textbf{LRP}-derived \textbf{R}OI \textbf{S}election Framework for ASD Diagnosis via Inter-regional High-level Relation Learning}
% \title{A Novel ROI Selection Framework for ASD Diagnosis via Inter-regional High-level Relation Learning}
\title{EAG-RS: A Novel Explainability-guided ROI-Selection Framework for ASD Diagnosis via Inter-regional Relation Learning}

\author{Wonsik Jung, Eunjin Jeon, Eunsong Kang, and Heung-Il Suk
\IEEEmembership{Member, IEEE}
\thanks{This research was supported by the National Research Foundation of Korea (NRF) grant funded by the Korea government (No. 2022R1A4A1033856) and the Institute of Information \& communications Technology Planning \& Evaluation (IITP) grant funded by the Korea government (MSIT) No. 2022-0-00959 ((Part 2) Few-Shot Learning of Causal Inference in Vision and Language for Decision Making) and (No. 2019-0-00079, Artificial Intelligence Graduate School Program(Korea University)).}
% \thanks{This paragraph of the first footnote will contain the date on which
% you submitted your paper for review. It will also contain support information,
% including sponsor and financial support acknowledgment. For example, 
% ``This work was supported in part by the U.S. Department of Commerce under Grant BS123456.'' }
\thanks{W. Jung, E. Jeon, and E. Kang are with the Department of Brain and Cognitive Engineering, Korea University, Seoul 02841, Republic of Korea (e-mail: \{ssikjeong1, eunjinjeon, eunsong1210\}@korea.ac.kr)}
% \thanks{Third Author is with the Department of Artificial Intelligence, Korea University, Seoul 02841, Republic of Korea (e-mail: \{-\}@korea.ac.kr)}
\thanks{H.-I. Suk is with the Department of Artificial Intelligence, Korea University, Seoul 02841, Republic of Korea and also with the Department of Brain and Cognitive Engineering, Korea University, Seoul 02841, Republic of Korea (e-mail: hisuk@korea.ac.kr).}
\thanks{Corresponding author: Heung-Il Suk}}
\maketitle

\begin{abstract}
Deep learning models based on resting-state functional magnetic resonance imaging (rs-fMRI) have been widely used to diagnose brain diseases, particularly autism spectrum disorder (ASD). Existing studies have leveraged the functional connectivity (FC) of rs-fMRI, achieving notable classification performance. However, they have significant limitations, including the lack of adequate information while using linear low-order FC as inputs to the model, not considering individual characteristics (\ie, different symptoms or varying stages of severity) among patients with ASD, and the non-explainability of the decision process. To cover these limitations, we propose a novel explainability-guided region of interest (ROI) selection (EAG-RS) framework that identifies non-linear high-order functional associations among brain regions by leveraging an explainable artificial intelligence technique and selects class-discriminative regions for brain disease identification.
The proposed framework includes three steps: (i) inter-regional relation learning to estimate non-linear relations through random seed-based network masking, (ii) explainable connection-wise relevance score estimation to explore high-order relations between functional connections, and (iii) non-linear high-order FC-based diagnosis-informative ROI selection and classifier learning to identify ASD. We validated the effectiveness of our proposed method by conducting experiments using the Autism Brain Imaging Database Exchange (ABIDE) dataset, demonstrating that the proposed method outperforms other comparative methods in terms of various evaluation metrics. 
Furthermore, we qualitatively analyzed the selected ROIs and identified ASD subtypes linked to previous neuroscientific studies.
\end{abstract}


\section{Introduction}
\label{sec:introduction}
\IEEEPARstart{A}{}utism spectrum disorder (ASD) is a neurological disability associated with brain development. Patients with ASD experience social communication and interaction difficulties in multiple contexts, and exhibit limited or repetitive behavioral patterns, interests, or activities \cite{ecker2010describing}. 
Although patients with ASD incur considerable average medical expenses over their lifetime (\eg, at least one million dollars per patient \cite{buescher2014costs}), accurate clinical curative treatments are not available, forcing them to suffer from lifelong illnesses \cite{fernell2013early}.
Therefore, it is crucial to identify the emergence of the disease as early as possible for accurate treatment \cite{zwaigenbaum2015early}.

Over the past few decades, several approaches based on resting-state functional magnetic resonance imaging (rs-fMRI) have been proposed to diagnose various brain diseases, including ASD \cite{abraham2017deriving}, schizophrenia \cite{kim2016deep}, and Alzheimer's disease \cite{brier2012loss}.
Rs-fMRI is a non-invasive technique that identifies spatio-temporal scales of regional brain activation by measuring blood-oxygen-level-dependent (BOLD) signals \cite{cribben2012dynamic}.
Most existing rs-fMRI studies utilize raw time signals \cite{dvornek2017identifying,kang2018probabilistic,jeon2020enriched} or low-order brain functional connectivity (FC)~\cite{chen2016high,zhao2020diagnosis}. FC is typically constructed based on the temporal correlation between spatially remote brain regions---regions of interest (ROIs)---in a statistical manner \cite{biswal2012resting,lee2003report}. Therefore, FC not only provides information about functional communication in the human brain \cite{van2010exploring} but also employs it as a biotype for the disease diagnosis~\cite{wager2017imaging}.

With recent advances in deep learning (DL), brain disease diagnostic methods based on rs-fMRI have garnered significant attention in neuroimaging research.
Raw time signals of rs-fMRI have been used as inputs in recurrent neural networks (RNNs) \cite{dvornek2017identifying}, graph convolutional neural networks (GCNs) \cite{parisot2018disease}, \etc, and FC has been used as input in a variant of auto-encoder (AE) architecture \cite{heinsfeld2018identification,suk2016state,guo2017diagnosing,eslami2019asd,wang2019identification,rakic2020improving,jung2021inter} to develop methods for diagnosing brain diseases.
In addition, several approaches have utilized more discriminative features for improving diagnostic performance~\cite{wee2012identification,tibshirani1996regression,guo2017diagnosing,wang2019identification}.
Feature selection (FS) methods involve (i) ranking-based methods, in which all features are ranked and then curated based on their specific criteria~\cite{wee2012identification,guo2017diagnosing}, and (ii) subset-based methods, which select features by optimizing a definite objective function~\cite{tibshirani1996regression,wang2019identification}.
Moreover, FS methods help explore the pathology of brain diseases by considering the selected features as biomarkers \cite{wang2019graph}.

Although these diagnostic methods exhibit remarkable classification performance, they continue to suffer from certain limitations. 
Firstly, most existing methods use (partial) Pearson's correlation as their input FC, which represents the linear correlation of brain regions as connectivity strength and contains low-order information within brain regions or voxels. However, {\rev merely considering} low-order information is not sufficient for capturing subtle changes in signal between normal and patient groups \cite{zhu2019hybrid}.
{\rev \cite{zhang2016topographical,zhao2021constructing} proposed a method for constructing high-order FC networks based on the similarity between the topographical profiles of pairs of FCs, which is referred to as ``correlation's correlation". In contrast, existing studies on FC typically calculate low-order or high-order FCs separately. However, pairs of FC levels may exhibit intriguing relationships or functional associations. In this context, \cite{zhang2017hybrid} integrated three types of FCs, encompassing low-order FC, high-order FC, and the inter-level associated FC, and proposed a hybrid high-order FC network for brain disease diagnosis tasks. Their method exhibited higher accuracy than methods based on a single type of FC. Based on the findings reported in prior studies, our study aims to enhance diagnostic performance by leveraging a combination of low-order FC and high-order features.}

Ranking-based FS approaches focus on single levels of contribution, and therefore, do not consider complementary information between multiple features \cite{wang2019graph}.
On the other hand, subset-based FS methods investigate the importance of various groups of features simultaneously and do not consider the individual characteristics of patients with ASD.
Finally, a few other FS methods use refined inputs and ignore local-global structural information in terms of the entire population, making their decision-making processes difficult to explain.
In practice, explainability is vital in the medical field (especially in neuroimaging) to improve reliability.

To address the aforementioned issues, we propose a novel explainability-guided ROI selection (EAG-RS) framework that selects informative features dynamically at the ROI-level for brain disease diagnosis.
To this end, we estimate high-order information of FC based on a high-level representation obtained from the layer-wise relevance maps. We leverage the estimated information in conjunction with low-order information for ASD diagnosis learning.
{\rev Further, prior studies~\cite{heinsfeld2018identification,rakic2020improving} have primarily focused on learning low-level inter-regional FC relationships based on computer vision tasks rather than from a neuroscientific perspective. Our earlier work~\cite{jung2021inter} introduced a novel approach from a neuroscientific perspective to complement these limitations, emphasizing brain region-level considerations. We designed and used random ROI-level masking to facilitate robust and expressive feature learning.}
Given an ROI-masked FC, a stacked AE (SAE) inherently learns non-linear relations among remaining ROI connections to reconstruct or infer masked connections. Following model training, connection-wise relevance score estimation is performed based on the {\rev pre-trained SAE} with layer-wise relevance propagation (LRP) \cite{bach2015pixel} to explore the high-order relations between functional connections.
The LRP transmits the output of the trained network back to the input level using a decomposition rule, which {\rev enables the identification of input connection features that contribute to the restoration of masked connections either positively or negatively.} 
Thus, we estimate non-linear high-order relations among seed-based networks (\ie, ROIs) in FC via the trained inter-regional non-linear relational learning model. 
Finally, given the estimated non-linear high-order FC, the pre-trained encoder and classifier are trained to discover ASD-informative ROIs at the sample level.

The proposed method is verified to exhibit superior classification performance than comparative methods on the publicly available Autism Brain Imaging Database Exchange (ABIDE) dataset \cite{di2014autism}.
The impacts of individual components of the proposed method are estimated using ablation studies and post-hoc analysis is performed to identify ASD subtypes.
The main contributions of our study can be summarized as follows:
{\begin{itemize}
\item {We propose a novel method to derive the high-order information of FC using a connection-wise relevance score estimation module between each masked seed ROI and other neighboring ROIs.}

\item {Two types of representative vectors (\ie, mean and count) statistically estimated based on the connection-wise relevance score, contribute to the selection of disease-relative ROIs at the individual level.}
            
\item Our proposed framework achieves state-of-the-art diagnosis performance on the ABIDE dataset. The neuroscientific analysis is also conducted using our framework.
\end{itemize}
}

This study is an extension of our previous work~\cite{jung2021inter}, in which we introduced a self-supervised learning framework that considers inter-regional non-linear relations for rs-fMRI. In this study, the proposed framework is supplemented using connection-wise relevance score estimation and a diagnosis-informative ROI selection network, thereby improving its ASD diagnosis performance. We further conducted an ablation study to verify the capabilities of the constituent modules in our proposed framework. In addition, we utilized selected ROIs at the sample level for cluster subtyping of autism and analyzed them to acquire neuroscientifically reliable results by performing group-wise ROI comparisons.

\begin{figure*}[!t]
\centerline{\includegraphics[width=1.7\columnwidth]{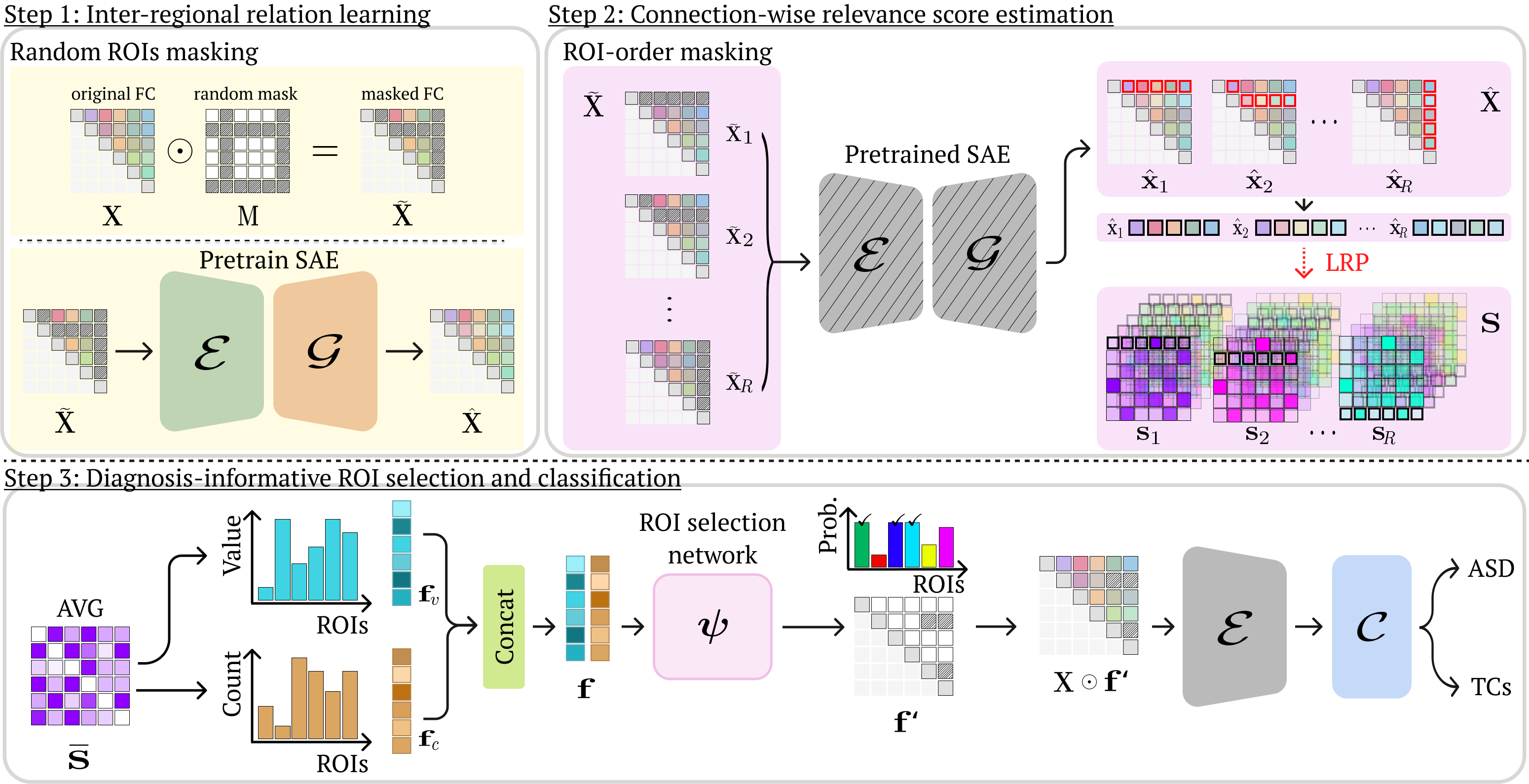}}
\caption{Overview of the EAG-RS framework comprising a three-step learning strategy: i) inter-regional relation learning using seed (ROI)-based network masking that generates masked input FCs for ROIs, ii) estimating connection-wise relevance scores via LRP to investigate high-order information between functional connections, and iii) extracting ROI-level representative vectors from the estimated relevance scores for simultaneous diagnosis-informative ROI selection and brain disease diagnosis. Slashed modules (\ie, pre-trained SAE) represent modules with frozen parameters (\ie, fixed parameters). {\revv Given a masked FC, the SAE is trained to discern non-linear relations among low-order FC connections. Henceforth, we estimate connection-wise relevance scores using the pre-trained SAE combined with LRP. These scores signify the importance of a given connection in restoring other connections. Lastly, we integrate low-order and high-order FCs to individually select informative ROIs for ASD diagnosis.}}
\label{fig1}
\end{figure*}

\section{Related Works}
\label{sec:related_work}
\subsection{DL-based Brain Disease Diagnosis Methods in rs-fMRI}
\label{subsec:ASDDiagnosis}
{In recent decades, DL has been utilized to diagnose brain diseases using rs-fMRI \cite{jeon2020enriched,dvornek2017identifying,parisot2018disease,heinsfeld2018identification,eslami2019asd,guo2017diagnosing,wang2019identification,liu2020improved}.
In \cite{dvornek2017identifying} and \cite{jeon2020enriched}, diagnosis tasks were performed using the raw time signals of rs-fMRI.
In particular, \cite{dvornek2017identifying} first trained a long short-term memory network (LSTM) to capture greater amounts of temporal information. 
\cite{jeon2020enriched} focused on extracting more disease-relevant intermediate features by combining a self-attention mechanism with mutual information maximization. Another approach is to use the FC of the rs-fMRI as the input for DNNs. For example, \cite{heinsfeld2018identification} and \cite{eslami2019asd} learned the hidden representations of FC by training AE variants and using them for classification.

Several studies have demonstrated that the FS-based methods enhance classification performances. \cite{guo2017diagnosing} proposed the DNN-FS method, in which the authors ranked outputs of multiple stacked sparse AEs in terms of their Fisher scores to determine more discriminative features. 
\cite{wang2019identification} devised support vector machine recursive feature elimination (SVM-RFECV). Further, \cite{liu2020improved} selected features based on a combination of elastic net and manifold regularization, referred to as MTFS-EM.

However, existing diagnosis methods result in suboptimal performance because of their single-level contribution to low-order information, without considering individual characteristics. 
In contrast, our proposed method presumes the high-order FC information dynamically and explores more discriminative features by incorporating estimations and explainable relevance maps.}

\subsection{Explainability for fMRI}
\label{subsec:LRP}
Deep neural networks (NNs) suffer from a black-box problem, wherein their outcomes cannot be easily explained because of complex non-linear mechanisms. To address this issue, various {\revv explainable artificial intelligence (XAI)} methods have been applied to neuroimaging models \cite{yan2017discriminating,azevedo2022deep,zhao2022attention,lin2022sspnet}. Among these, LRP is widely used to identify group-discriminative features/patterns by quantifying their relevance to the model outcome \cite{jung2021inter,yan2017discriminating,zhao2022attention}. For example, \cite{zhao2022attention} analyzed disease classification results by searching for the most group-discriminative patterns using LRP.

{\rev Besides providing neuroscientific explanations,} LRP can {\rev also be used to construct novel FC-determination techniques.} For example, a novel brain connectivity measurement based on a trained network with BOLD signals was reported using LRP~\cite{dang2019novel}.
The objective of this paper was to explain the relevance of one region in influencing another in a regression task using NNs.

Similar to \cite{dang2019novel}, we explore brain connectivity using LRP in this paper. The main differences between \cite{dang2019novel} and our study are as follows:
(i) We use a seed-based network mask as the input of the FC network, instead of BOLD signals, which include self-seed-ROI influences. Therefore, we focus on the inter-regional non-linear relationships to estimate the contributions of neighbors inherently restoring a seed-based network.
(ii) We apply LRP results to explore high-order relations between functional connections using connection-wise relevance score estimation and leverage them to select ASD-discriminative ROIs dynamically during training. This is unlike any technique in any previously published LRP-based study.

\section{Methods}
The overall framework of the proposed EAG-RS is illustrated in Fig~\ref{fig1}. It comprises three phases; i) inter-regional relation learning, ii) connection-wise relevance score estimation via LRP, and iii) brain disease diagnosis based on diagnosis-informative ROI selection.
Given an FC randomly masked at the seed-based network level, the SAE is trained to learn a feature representation that reflects non-linear relations between low-order FC connections.
The proposed framework estimates the connection-wise relevance score, which represents the relevance of a connection to the restoration of other connections using the trained SAE model and LRP. This framework helps analyze FCs between spatially distinct regions. Then, ASD-informative ROIs are selected at an individual level based on statistical measures of the relevance scores (\ie, mean and count). Finally, we diagnose ASD by considering only ASD-informative ROIs in FC.

\subsection{ROI connection masking}
\label{subsubsec:ROImasking}
%\subsubsection*{ROI connection masking}

Given an input FC matrix $\bX\in\mathbb{R}^{R\times R}$, where $R$ denotes the number of ROIs, we generate {\rev a mask matrix $\mathbf{M}$ at the ROI-level, where the set $\mathbf{R}_q$ includes randomly selected ROI indexes based on a $q$-ratio. The} mask matrix $\mathbf{M}\in\boldsymbol{1}^{R\times R}$ is updated using the following rule: {\rev $\mathbf{M}(i,:)=0$ and $\mathbf{M}(:,i)=0$ for $i\in\mathbf{R}_q$,} where $(i,:)$ and $(:,i)$ denote the elements of the $i$-th row and any column, and any row and the $i$-th column, respectively.
{\rev Then, the masked input FC matrix $\tilde{\mathbf{X}}$ is obtained via element-wise multiplication of the input FC matrix $\mathbf{X}$ with the mask matrix $\mathbf{M}$, and it is denoted by $\tilde{\mathbf{X}}=\mathbf{M}\odot\mathbf{X}$}, where $\odot$ represents an element-wise multiplication operator.
{\rev The masked FC matrix $\tilde{\mathbf{X}}$ is flattened to a one-dimensional vector} $\tilde{\mathbf{X}}\in\mathbb{R}^{D}$, where $D$ indicates the number of elements in the upper triangle of the FC matrix without diagonal elements ($D = R\times(R-1)/2$).
In this procedure, masks are generated arbitrarily for each input during each iteration. {\rev This enables augmentation of the training samples and helps in learning robust and enriched feature representations, thereby preventing overfitting~\cite{he2022masked}.}

\subsection{Inter-regional relation learning}

In contrast to our previous work~\cite{jung2021inter}, in this paper, we focus on inter-regional relation learning without considering classification to be the first step for examining high-order information of FCs. 
Initially, the flatten-masked FC $\tilde{\mathbf{X}}$ is embedded into a hidden space with $\mathbf{h}_1\in\mathbb{R}^{D_1}$, as follows:
\begin{equation}
\mathbf{h}_1 = \mathcal{E}_1(\tilde{\mathbf{X}})=\sigma(\mathbf{W}_1\tilde{\mathbf{X}}+\mathbf{b}_1),
\end{equation}
where $\mathcal{E}_1$ denotes the first layer of the encoder with a weight matrix $\mathbf{W}_1\in\mathbb{R}^{D_1\times D}$ and a bias vector $\mathbf{b}_1\in\mathbb{R}^{D_1}$; and $\sigma$ denotes the activation function. The first hidden represented features $\mathbf{h}_1$ is trained to reconstruct the original FCs $\mathbf{X}$ using the corresponding layer of the decoder (\ie, generator) 
$\mathcal{G}_1(\mathbf{h}_1)=\tanh(\mathbf{W}'_1\mathbf{h}_1+\mathbf{b}'_1)=\hat{\mathbf{X}}$ by minimizing the reconstruction loss, as {\rev follows:}
\begin{equation}
\label{eq:loss_rec1}
\min_{\mathbf{W}_1,\mathbf{W}'_1,\mathbf{b}_1,\mathbf{b}'_1}\mathcal{L}_{rec}(\mathbf{X},\hat{\mathbf{X}})=\sum_{i=1}^{N}||\mathbf{X}^{(i)}-\mathcal{G}_1(\mathbf{h}_1^{(i)})||,
\end{equation}
where $N$ and $\hat{\mathbf{X}}$ represent the total numbers of training and reconstructed samples, respectively, which are outputs of $\mathcal{G}_1$; and $\mathbf{W}'_1\in\mathbb{R}^{D\times D_1}$ and $\mathbf{b}'_1\in\mathbb{R}^D$ denote a weight matrix and a bias vector, respectively. Note that ${\Theta}_{\mathcal{E}}$ and ${\Theta}_{\mathcal{G}}$ are learnable parameters; thus, $\{\mathbf{W}_1,\mathbf{b}_1\}\subset{\Theta}_{\mathcal{E}}$,$\{\mathbf{W}_1',\mathbf{b}_1'\}\subset{\Theta}_{\mathcal{G}}$.

{\rev During the training process, to learn high-level feature representations of FCs, we sequentially} train the encoder ($\mathcal{E}_\ell$) and generator ($\mathcal{G}_\ell$) pair {\rev for each layer $\ell\in\{2, \dots, L\}$ in the network.} {\rev We freeze} the previous layer(s) of both the encoder and the generator {\rev while estimating} the $\ell$-th level representation of the FC input, $\mathbf{h}_\ell$, which corresponds to the $(\ell+1)$-th level non-linear relations among ROIs. {\rev Subsequently, this is transmitted to} $\mathcal{E}_\ell$.
The encoder $\mathcal{E}_\ell$ and the subsequent generator $\mathcal{G}_\ell(\mathbf{h}_\ell)=\tanh(\mathbf{W}'_\ell\mathbf{h}_\ell+\mathbf{b}'_\ell)=\hat{\mathbf{h}}_{\ell-1}$ are trained by minimizing the sum of the reconstruction losses of $\{\mathbf{h}_\ell\}_\ell^L$ and $\mathbf{X}$.
In this regard, the proposed SAE is trained to reconstruct the removed connections as well as estimate the remaining connections based on relations inherently present in the neighboring connections.
{\rev In this step}, the proposed model learns {\rev an} inter-regional non-linear representation {\rev that encompasses} first-order connections {\rev of} rs-fMRI {\rev as well as} high-level relations among ROIs.

\subsection{Connection-wise relevance score estimation}

{\rev After training the SAE using the process stated in Section III.B, w}e utilize the LRP technique to estimate connection-wise relevance scores in {\rev the pre-trained} SAE.
{\rev The relevance score represents the influence of each connection on other connections. 
The LRP traces back} from the final output layer to the input connection layer {\rev to calculate these scores}.
{\rev Specifically, we} define the relevance score $S_j^{\ell+1}$, which represents a hidden unit $j$ in the $(\ell+1)$-th layer. {\rev The relevance score $S_j^{\ell+1}$ is determined based on the contribution of} all hidden units in {\rev the $\ell$-th layer} that {\rev affect} the activation of {\rev the} hidden unit $j$ in the subsequent layer $\ell+1$. {\rev This ensures} that the total relevance per layer is conserved~[29] as {\rev $\sum_i 
S_{i\leftarrow j}^{\ell,\ell+1}=S_j^{\ell+1}$.}

Given the original FC {\rev matrix} ${\mathbf{X}}$, {\rev which includes a} set of $R$ seed-based networks, {\rev we generate a masked FC} $\tilde{\mathbf{X}}$ {\rev by removing one of the seed-based networks (\ie, ROI), resulting in a set of $(R-1)$ seed-based networks. To achieve this,} the $r$-th seed-based networks are masked in the sequence of ROI indexes.
{\rev Subsequently,} the masked FC matrix is transmitted to the pre-trained SAE {\rev to reconstruct the original FC matrix, denoted by $\hat{\mathbf{X}}$. Via this reconstruction process, the masked FC reconstructs the masked seed-based network based on the remaining $(R-1)$ non-masked seed-based networks, as follows:}
\begin{equation}
	\hat{\mathbf{X}} = \text{pre-trained SAE}(\tilde{\mathbf{X}})=\mathcal{G}(\mathcal{E}(\tilde{\mathbf{X}})),
\end{equation} where $\mathcal{E}$ and $\mathcal{G}$ represent the encoding and decoding layers {\rev of the pre-trained} SAE, respectively.
% In contrast to randomly masking $q\%$ seed-based networks for training an SAE, our pre-trained SAE reconstructs one missing seed-based network from the other $(R-1)$ non-masked seed-based networks. 
{\rev This process is repeated $R$ times for each seed-based network in the FC, resulting in} $\hat{\mathbf{X}}=\{\hat{\mathbf{x}}_{r}^{\top}\}_{r=1,\ldots,R}$.
{\rev Subsequently, we use the LRP technique to estimate the relevance scores, representing} the contributions of other connections to the masked seed-based networks.

The reconstructed FC $\hat{\mathbf{X}}=\{\hat{\mathbf{x}}^\top_1,\ldots,\hat{\mathbf{x}}^\top_r,\ldots,\hat{\mathbf{x}}^\top_R\}$ {\rev is utilized to estimate} the connection-wise relevance score $\mathbf{S}=\{\mathbf{s}_1,\ldots,\mathbf{s}_r,\ldots,\mathbf{s}_R\}$ via LRP.
{\rev The relevance score represents the contributions of non-masked neighboring ROIs in restoring masked seed-ROI connections.}
To obtain this, we define the $\phi(\cdot)$ function, {\rev which assigns a value of 0 to masked regions (\ie, the $i$-th row) and applies LRP to the remaining non-masked regions, as follows}
\begin{equation}
% \hat{\mathbf{x}}_r =  \sum_{j=1}^{R}\phi(\hat{{x}}_{r,j},j),~
\phi(\hat{\mathbf{x}}^\top_{i,j},i,j)=\begin{cases}
\boldsymbol{0}& \text{if}~i\text{-th}~\text{row},\\
\text{LRP}(\hat{\mathbf{x}}^\top_{i,j})&
\text{otherwise}
\end{cases}
\label{eq:lrp1}
\end{equation}
where $i$ and $j$ represent the corresponding ROI indexes and $\dim(\boldsymbol{0})=1\times R$, respectively.
{\rev The original dimension of the LRP outcomes, obtained when the entire FC matrix is provided as input to the pre-trained SAE without masking, is $R\times R$}.
However, {\rev since we mask} the $i$-th seed-based networks {\rev before transmitting them to} the pre-trained SAE, the masked regions in the {\rev LRP outcomes} can be disregarded. Therefore, the dimension of $\text{LRP}(\hat{\mathbf{x}}_{i,j}^\top)$ in {\rev Eq. \eqref{eq:lrp1}} is $(R-1)\times R$.

{\rev The connection-wise relevance score $\mathbf{s}_r$, where $r\in\{1,\ldots,R\}$ for reconstructing the $r$-th ROI based on other connections, is obtained using the $\phi(\cdot)$ function and defined as}
\begin{equation}
\mathbf{s}_{r} = 
\bigg\Vert_{j=1}^{R}{\phi(\hat{\mathbf{x}}^\top_{{r},j},{r},j)},
% \mathbf{s}_{r} = 
% \underbrace{\bigg\Vert_{j=1}^{R}\underbrace{\phi(\hat{\mathbf{x}}^\top_{r,j},r,j)}_{\mathbf{s}_{r,j}\in\mathbb{R}^{R\times R}}}_{\mathbf{s}_r\in\mathbb{R}^{R\times R\times R}},
\label{eq:lrp2}
\end{equation}
where $\bigg\Vert$ indicates a concatenation operator {\rev and dim($\mathbf{s}_{r})=R\times R\times R$}.
% , the relevance score for contributions to reconstructing $r$-th ROI from other connections, respectively. 
Note that the self-connections corresponding to the seed-ROI are {\rev excluded during the calculation of} the relevance score. 
% {\rev In other words}, $\mathbf{s}_r$ represents the local contributions of $r$-th ROI to several brain regions. 
{\rev To estimate the local contributions of the $r$-th ROI, we simply aggregate the relevance scores for various connections~\cite{gholizadeh2021model}. This can be done using the following equation} 
\begin{equation}
    \mathbf{s}'_{\rev r} = {\rev \sum^{R}_{{k}=1}\mathbf{s}(\cdot,\cdot,k)},
    \label{eq:lrp3}
\end{equation}
where $\mathbf{s}'_r\in\mathbb{R}^{R\times R}$ denotes the aggregated relevance score.
Moreover, {\rev a global explanation can be represented} by aggregating all the connections from the perspective of the $r$-th ROI.

By increasing the order of ROI indexes $r$ from $1$ to $R$ as given by {\rev Eq.~(\ref{eq:lrp2}) and (\ref{eq:lrp3})}, we obtain the global explanation set $\mathbf{S}$, as follows: 
\begin{align}
	\mathbf{S} &= \bigg\Vert^{R}_{r=1}\mathbf{s}'_r.
\end{align}
{\rev The detailed procedure is outlined} in Algorithm~1.
Finally, the ROI selection network takes the mean of the set $\mathbf{S}$ as the input.

\begin{algorithm}[t!]
\caption{Connection-wise relevance score estimation}
\label{algo2}
\SetAlgoLined
\SetKwInOut{Input}{input}
\SetKwInOut{Output}{output}
\DontPrintSemicolon
	%	\Input{Training dataset $\mathbf{X}, \mathbf{Y}$; network architectures $\mathcal{E}, \mathcal{G}$; network parameters $\Theta_{\mathcal{E}}, \Theta_{\mathcal{D}}$; a stochastic gradient descent optimizer SGD and its hyperparameter set $\boldsymbol{\eta}$}
	\Input{Masked FC $\tilde{\mathbf{X}}$}
	\Output{Relevance score $\mathbf{S}$}
	Define a set $A=\{1,2,\ldots,R\}$ \textit{\# ROI indices}
	
	Initialize $\mathbf{S}=[]$	
	
	\For{$r=1,\ldots,R$}{
		Draw data $\tilde{\mathbf{X}}$ containing masked $r$-th ROI indexes
		
		$\hat{\mathbf{X}}\leftarrow\text{pre-trained SAE}(\tilde{\mathbf{X}})$
		
		\For{$j=1,\ldots,R$}{
			\eIf{$j=r$}{
				$\mathbf{s}_r$.append($\boldsymbol{0}$) \textit{\# $\dim(\boldsymbol{0})=R$}\Comment{Eq. (5)}
				%				\textit{continue}
				%				\textit{\# excluding self-connection}
			}{
				$A\leftarrow A-\{r\}$ \textit{\# excluding self-connection}
				
				$\mathbf{s}_{r}$.append(LRP($\hat{\mathbf{X}}^\top_{r,j}$)[$A$,:]) \Comment{Eq. (5)}
				
				\textit{\#} $\dim(\text{LRP}(\hat{\mathbf{X}}^\top_{r,j}))=R\times R$
			}
			$\mathbf{s}'_{r}\leftarrow$ np.sum($\mathbf{s}_{r}$, dim=2)\Comment{Eq (6)}
   
            \textit{\# $\dim(\mathbf{s}'_{r})=R\times R$}
			
			%			\textit{\# adding the contribtuions}
			%			\eIf{$j=r$}{
			%%				$\mathbf{s}_r.append(\boldsymbol{0}$) 
			%				\textit{continue}
			
			%				\textit{\# excluding self-connection}}
			%			{LRP($\tilde{\mathbf{x}}_k$)}
		}
		$\mathbf{S}$.append($\mathbf{s}'_{r}$) \textit{\# $\dim(\mathbf{S})=R\times R\times R$}\Comment{Eq (7)}
	}
	
\end{algorithm}

\subsection{ROI selection network and diagnostic classifier}
\subsubsection{ROI selection network ($\psi$)}
{\rev We perform statistical analysis to distinguish individual impacts and identify the most important effects~\cite{gholizadeh2021model}}. Therefore, given the averaged relevance scores $\bar{\mathbf{S}}\in\mathbb{R}^{R\times R}$, we reformulate the ROI-level representative vectors (\ie, $\mathbf{f}_v\in\mathbb{R}^{R\times 1}$ and $\mathbf{f}_c\in\mathbb{R}^{R\times 1}$) {\rev using statistical measures} such as mean and count, as referenced in Algorithm~\ref{algo3}.

Subsequently, these vectors are concatenated channel-wise as $\mathbf{f}=[\mathbf{f}_v\Vert\mathbf{f}_c]\in\mathbb{R}^{R\times 2}$, and transmitted into the ROI selection network $\psi$. The joint training of the ROI selection network and classifier reveals discriminative features for diagnosis.
To maintain the information {\rev corresponding to} each ROI, we use a convolutional layer (Conv1D) with a learnable $2\times 1$ kernel, a stride of one in each dimension, and zero padding. Based on these configurations, we define the ROI selection network as follows:
\begin{align}
\hat{\mathbf{f}}&=\psi(\mathbf{f})\\
&=\text{Gumbel-softmax}(\mathbf{W}_{\psi_2}\sigma(\mathbf{W}_{\psi_1}\mathbf{W}_{\text{Conv1D}}(\mathbf{f})+\mathbf{b}_{\psi_1})+\mathbf{b}_{\psi_2})\nonumber
\end{align} where $\mathbf{W}_{\psi_1}$, $\mathbf{W}_{\psi_2}$, and $\mathbf{W}_{\text{Conv1D}}$ denote weight matrices, $\mathbf{b}_{\psi_1}$ and $\mathbf{b}_{\psi_2}$ denote bias vectors, $\sigma$ denotes a Rectified Linear Unit (ReLU) activation function, and $\Theta_\psi$ denotes a learnable parameter. 
% Thus, $\{\mathbf{W}_{\psi_1},\mathbf{W}_{\psi_2},\mathbf{b}_{\psi_1},\mathbf{b}_{\psi_2},\mathbf{W}_{\text{Conv1D}}\}$.

\begin{algorithm}[t!]
\caption{Formulating ROI-level vectors}
\label{algo3}
\SetAlgoLined
\SetKwInOut{Input}{input}
\SetKwInOut{Output}{output}
\DontPrintSemicolon
%	\Input{Training dataset $\mathbf{X}, \mathbf{Y}$; network architectures $\mathcal{E}, \mathcal{G}$; network parameters $\Theta_{\mathcal{E}}, \Theta_{\mathcal{D}}$; a stochastic gradient descent optimizer SGD and its hyperparameter set $\boldsymbol{\eta}$}
\Input{Dataset $\{\mathbf{X},\mathbf{S},\mathbf{Y}\}$}
\Output{ROI-level representative vectors $\mathbf{f}_v, \mathbf{f}_c$}

%	Initialize a mask matrix $\mathbf{M}=\mathbf{J}_R$ where $m_{(i,j)}=\phi(R_q)$ $(i,j\in R_q, R_q\subset R$ and $R_q\ne\emptyset)$
%	Initialize a mask matrix $\mathbf{M}$ where $m_{(i,j)}=\phi(R_q)$ $(i,j\in R_q, R_q\subset R$ and $R_q\ne\emptyset)$

%	//Initialization
%	
%	\quad \quad \quad $PS^* \gets \emptyset$
%	// Step 1: Inter-regional relation learning
$\bar{\mathbf{S}}=\text{np.mean}(\mathbf{S}, \text{dim=2})$ \textit{\# $\dim(\bar{\mathbf{S}})=(N\times R\times R$)}

Initialize a temporary mask $\mathbf{K}$ \textit{\# $\dim(\mathbf{K})=(N\times R \times R$)}

Initialize $\mathbf{f}_v,\mathbf{f}_c=[],[]$

\For{$n=1,\ldots,N$}{
    \For{$r=1,\ldots,R$}{
        
        \eIf{$\bar{\mathbf{s}}^n_r<$ \normalfont{np.mean}($\bar{\mathbf{s}}^n_r$)}{
            $\mathbf{k}^n_r\leftarrow \boldsymbol{0}$}
        {$\mathbf{k}^n_r\leftarrow \boldsymbol{1}$}
    }
    
    $\bar{\mathbf{s}}^n_r\leftarrow\bar{\mathbf{s}}^n_r\odot\mathbf{k}^n_r$
    
    $\mathbf{f}_v$.append(np.sum($\bar{\mathbf{s}}^n_r$, dim=2))
    
    $\mathbf{f}_c$.append(np.sum($\mathbf{k}^n_r$, dim=2))
}
\end{algorithm}

\subsubsection{Diagnostic classifier ($\mathcal{C})$}
{Information of individually selected ROIs, $\hat{\mathbf{f}}\in\mathbb{R}^{R\times 1}$, is reshaped and multiplied with the original FC $\mathbf{X}$.
{\rev To this end,} we perform the following operation: $\hat{\mathbf{f}}=\hat{\mathbf{f}}\odot\boldsymbol{1}^\top$, where $\boldsymbol{1}^\top\in\mathbb{R}^{1\times R}$ represents a vector of size $R$ containing a single value.}
In addition, to reflect the symmetrical characteristics of FCs, we perform the following operation:
\begin{align}
	\mathbf{f}'&=(\hat{\mathbf{f}}+\hat{\mathbf{f}}^{\top})\odot\frac{1}{2}\mathbf{I},
\end{align} where $\mathbf{I}$ denotes the identity matrix.
Subsequently, the original FC $\mathbf{X}$ and information of individually selected ROIs  ($\mathbf{f}'\in\mathbb{R}^{R\times R}$) are element-wise multiplied and simultaneously transmitted to the encoders ($\mathcal{E}$) of the pre-trained SAE and the prediction network ($\mathcal{C}$) for the brain disease diagnosis task.
{Note that we remove the bias vector corresponding to the encoder layers to retain connections of zero values {\rev and prevent it from affecting} other connections.}
The diagnostic classifier is trained to predict the clinical status $\hat{\mathbf{y}}$ by minimizing cross-entropy loss, as follows:
\begin{align}
	\mathcal{L}_{\text{cls}}(\mathbf{y},\hat{\mathbf{y}})=-\frac{1}{N}\sum_{n=1}^N\mathbf{y}^n\cdot\log(\hat{\mathbf{y}}^n).
\end{align} where $N$ and $\mathbf{y}$ denote the numbers of training samples and class labels, respectively.

\subsection{Optimization}
The objective function {\rev corresponding to} each step {\rev comprises} different losses, and it is given by
\begin{align*}
	\text{Step 1}&:
	 \min_{\Theta_{\mathcal{E}},\Theta_{\mathcal{G}}}\alpha\mathcal{L}_{\text{rec}}(\bX,\hat{\bX})+(1-\alpha)\sum_{\ell=2}^{L}\mathcal{L}_{\text{rec}}(\mathbf{h}_{\ell-1},\hat{\mathbf{h}}_{\ell-1}),\\
	\text{Step 3}&: \min_{\Theta_{\psi}\Theta_{\mathcal{E}},\Theta_{\mathcal{C}}}\mathcal{L}_{\text{cls}}(\mathbf{y},\hat{\mathbf{y}}),
\end{align*}
{where $\alpha$ is a hyperparameter {\rev used to control} the ratio between two losses. The parameters of the encoder and generators are optimized by minimizing the combination of the reconstruction losses in Step 1. {\rev None of the parameters are updated during relevance score estimation via LRP in Step 2.} In Step 3, the proposed model performs a classification task. {\rev To this end,} we use a cross-entropy loss to train the pre-trained encoder of SAE, an ROI selection network, and a classifier.  
}

\section{Experiments}
\subsection{Dataset \& Pre-processing}
\label{subsec:dataset}
We use pre-processed rs-fMRI data collected from the publicly available ABIDE\footnote{\url{http://fcon_1000.projects.nitrc.org/indi/abide/}} dataset~\cite{di2014autism}. The ABIDE dataset includes previously collected structural MRI, rs-fMRI, and phenotypic data for use by the broader scientific community. It consists of $1,112$ subjects, including $539$ from individuals with ASD and $573$ corresponding to typical development (TD) (ages 7--64 years, median $14.7$ years across groups) from $17$ international sites\footnote{\{UM, NYU, MAX MUN, OHSU, SBL, OLIN, SDSU, CALTECH, TRINITY, YALE, PITT, LEUVEN, UCLA, USM, STANFORD, CMU, and KKI\}}. The ROIs fMRI series of all sites are downloaded from the pre-processed ABIDE dataset with a configurable pipeline for the analysis of connectomes (CPAC), band-pass filtering ($0.01-0.1\text{Hz}$), and no global signal regression, and {\rev it is parcellated using the} Harvard-Oxford (HO) atlas. After downloading the pre-processed data, $110$ ROIs are acquired using the HO atlas. {\rev At this stage}, samples with missing filenames and incomplete data {\rev are excluded; and the remaining} $880$ samples across $17$ international sites are utilized, which include $418$ ASD subjects and $478$ TD subjects. The Pearson correlation coefficient {\rev is used} to estimate FC.

\subsection{Experimental Settings}
To ensure a fair comparison, stratified five-fold cross-validation is conducted, where one fold is used for the validation set, another for the test set, and the remaining folds for the training set comprising all samples in the ABIDE dataset. Average performance is estimated in terms of the area under the receiver operating characteristic curve (AUC), accuracy (ACC), sensitivity (SEN), and specificity (SPEC). All proposed methods as well as competing methods are implemented using PyTorch and trained using a Titan RTX GPU on Ubuntu 18.04. All codes used in the experiments are available in a repository\footnote{\url{https://github.com/ku-milab/EAG-RS}}.

\subsubsection{Training Settings}
In the proposed SAE architecture, the encoder $\mathcal{E}$ comprises two fully-connected layers ($L=2$) with the units of $\{9000, 1800\}$.
The generator, $\mathcal{G}$, comprises two fully-connected layers with a reverse number of hidden units from the encoder. For the non-linear activation function ($\sigma$), the scaled exponential linear unit (SELU) is used for only the first intermediate layer in the encoder, and hyperbolic tangent (Tanh) is used for the remaining layers. The diagnosis-informative ROI selection network $\psi$ comprises one Conv1D layer and three fully-connected layers with units of $\{512,1650,110\}$. The classifier $\mathcal{C}$ {\rev comprises} two fully-connected layers with $\{10,2\}$ hidden units. The ReLU activation function is used for all intermediate layers. In the meantime, we set the sigmoid and softmax functions as the activation functions of the last layers of $\psi$ and $\mathcal{C}$, respectively.

In Step 1, $10\%$ of the ROIs ($q=0.1$) are randomly masked during every training iteration and the SAE is trained using Adam optimizer \cite{kingma2014adam} with a learning rate of $10^{-3}$ and a mini-batch size of $50$ over $300$ epochs. In addition, $\ell_2$ regularization is applied with a coefficient of $5\times10^{-5}$. We set $\alpha$ to be $0.5$. 
All trainable parameters in Step 3 are optimized using the same settings except for the learning rate ($10^{-4}$). The Gumbel-softmax temperature is set to $0.01$. Note that we take a grid search strategy for hyperparameter selection and select the best parameters based on the validation set results.

\subsubsection{Competing Methods}
{\rev The following six comparative methods are considered to evaluate the proposed method.} First, a basic AE, dAE \cite{vincent2008extracting}, and SAE are trained without any masking methods; they share the same architecture as that of EAG-RS. Further, EAG-RS is compared with the AE with $\mathbf{M}$, and SAE with Gaussian noise \cite{vincent2010stacked}. Henceforth, we denote these two baselines by AE (M) and SAE (G). To validate the effectiveness of ROI-level masking, the SAE is trained using random FC connection masking, SAE (FC-M), inspired by \cite{pathak2016context}. Additionally, we demonstrate the statistical significance between our proposed EAG-RS and competing methods {\rev based on} McNemar's test~\cite{mcnemar1947note}. We also compare EAG-RS with other simple feature selection methods{\rev, including ranking-based approaches, such as the $t$-test ($p<0.05$)~\cite{wee2012identification} and recursive feature elimination (RFE)~\cite{guyon2002gene}, as well as the} subset-based approach LASSO~\cite{tibshirani1996regression}. In the case of these {\rev three} methods, we utilize a linear SVM, which is a commonly used classifier in brain disease diagnosis~\cite{wang2019identification}. Here, we adopt the hyperparameter $C$ for SVM and $\lambda$ for LASSO in the sets of $\{10^{-3},10^{-2},\ldots,10^{3}\}$ and $\{0.001,0.002,\ldots,0.01\}$, respectively. {\rev In the case of RFE-SVM, RFE iteratively assigns SVM weights to each feature based on its importance to the brain disease diagnosis by eliminating the least informative and redundant features. Further, ASD-DiagNet~\cite{eslami2019asd}, which is a state-of-the-art approach that employs an autoencoder-based architecture with joint single-layer perception (SLP) training, is also considered. For feature selection in ASD-DiagNet, the 1/4 largest and 1/4 smallest Pearson's correlation values are used as input features based on the training data. The hyperparameter ranges for our experiments are derived from the values reported in~\cite{eslami2019asd}.

In addition, we re-implement and compare the results of the state-of-the-art methods. First, we select BrainNetCNN~\cite{kawahara2017brainnetcnn}, a convolutional neural network (CNN)-based model comprising edge-to-edge, edge-to-node, and node-to-graph convolutional filters, thereby utilizing the topological locality of brain network structures. Next, BrainGNN~\cite{li2021braingnn} is considered, which uses FC as a node feature and selects the top 10\% positive partial correlations as edge features. The architecture of BrainGNN consists of ROI-aware graph convolutional layers and ROI-selection pooling layers, along with a regularization loss term that softens the distribution of the node pooling scores, facilitating the prediction of neurological biomarkers. Finally, BrainNetTF~\cite{kan2022brain} is also considered. It exhibits a transformer-based architecture with an orthonormal clustering readout function that accounts for the similarity of ROIs within functional modules underlying brain regions. The hyperparameter configurations reported in our manuscript are adopted for each comparative method.}

\begin{table}[t!]
\caption{Averaged classification performance over ASD and TD. (FC: functional connectivity, FC-M: random FC connection mask, G: Gaussian noise, M: random seed-based network mask, $^*$: `$p<0.05$).}
\label{table1}
\centering{
\setlength{\tabcolsep}{2pt}
\scalebox{0.9}{
\begin{tabular}{lcccccc}\\\toprule
Models & \multirow{1}{*}{AUC} & \multirow{1}{*}{ACC (\%)} & \multirow{1}{*}{SEN (\%)} & \multirow{1}{*}{SPEC (\%)}\\\toprule
AE &0.676$\pm$0.043$^*$ & 62.11$\pm$3.63$^*$ & 59.41$\pm$7.26$^*$ & 64.67$\pm$2.42$^*$\\
AE (M)& 0.685$\pm$0.034$^*$  & 62.78$\pm$2.56$^*$  & 54.71$\pm$11.50$^*$ & 70.47$\pm$11.07$^*$\\
dAE~\cite{vincent2008extracting} & 0.678$\pm$0.044$^*$ & 62.58$\pm$3.95$^*$ & 59.22$\pm$7.32$^*$ & 65.79$\pm$4.70$^*$\\
{SAE}& 0.691$\pm$0.053$^*$ & 63.64$\pm$3.35$^*$ & 59.80$\pm$4.90 & 67.29$\pm$8.67\\
SAE (G)~\cite{vincent2010stacked}& 0.713$\pm$0.029$^*$  & 65.36$\pm$2.34$^*$  & 61.32$\pm$6.34 & 69.85$\pm$6.69\\
SAE (FC-M)& 0.735$\pm$0.033$^*$  & 66.35$\pm$3.70$^*$  & 57.19$\pm$5.53$^*$ & 74.65$\pm$7.54\\\midrule
\multirow{1}{*}{\textbf{Baseline}~\cite{jung2021inter}}& {0.757$\pm$0.040} & {69.76$\pm$3.45} & {57.82$\pm$6.64} & {80.22$\pm$4.54}\\
\multirow{1}{*}{\textbf{EAG-RS}}& \textbf{0.760$\pm$0.033} & \textbf{73.71$\pm$2.83} & \textbf{64.56$\pm$8.36} & \textbf{80.74$\pm$8.28}\\\bottomrule
\end{tabular}}
}
\end{table}

\begin{table}[]
\caption{Comparison of feature selection methods: Averaged classification performance.}
\label{table3}
\centering{
\setlength{\tabcolsep}{2pt}
\scalebox{0.9}{
\begin{tabular}{ccccccc}\\\toprule
    Methods &\multirow{1}{*}{AUC} & \multirow{1}{*}{ACC (\%)} & \multirow{1}{*}{SEN (\%)} & \multirow{1}{*}{SPEC (\%)}\\\toprule
    $t$-test+SVM & 0.705$\pm$0.030 & {65.59$\pm$3.47} & {63.82$\pm$5.76} & {67.03$\pm$5.02}\\
    LASSO+SVM & {0.727$\pm$0.030} & {66.88$\pm$3.80} & {63.01$\pm$5.18} & {70.93$\pm$4.95}\\
    {\rev RFE+SVM} & {\rev 0.706$\pm$0.010} & {\rev 65.00$\pm$2.25} & {\rev 63.04$\pm$1.09} & {\rev 67.60$\pm$1.45}\\
    {\rev ASD-DiagNet~\cite{eslami2019asd}}& {\rev 0.743$\pm$0.046} & {\rev 67.84$\pm$4.22} & {\rev 60.04$\pm$6.86} & {\rev 74.28$\pm$4.63}\\\midrule
    EAG-RS w/o $\psi$ & {0.757$\pm$0.040} & {69.76$\pm$3.45} & {57.82$\pm$6.64} & {80.22$\pm$4.54}\\
    \textbf{EAG-RS} &\textbf{0.760$\pm$0.033} & \textbf{73.71$\pm$2.83} & \textbf{64.56$\pm$8.36} & \textbf{80.74$\pm$8.28}\\
    \bottomrule
\end{tabular}}}
\end{table}

\subsection{Experimental Results}
\label{subsec:classification}
\subsubsection{AE-based classification results}
The comparative methods, such as AE, AE (M), and dAE, use different masking methods and are trained in an end-to-end manner. On the other hand, the SAE-based methods adopt greedy layer-wise training strategies~\cite{bengio2006greedy}, and their masking configurations are different from those of the AE-based methods. Experimental results are summarized in Table~\ref{table1}. {\rev SAE without a mask (SAE) is observed to outperform AE-based methods in terms of all metrics except for specificity.}
Meanwhile, SAE (G) outperforms SAE in terms of all metrics. In addition, SAE (FC-M) outperforms all competing methods in terms of AUC, ACC, and SPEC, but not sensitivity (SEN). 
Importantly, SAE with a random seed-based network mask (M), which was proposed in our previous work~\cite{jung2021inter} and is referred to as Baseline in this paper, is observed to outperform SAE (FC-M) in terms of all metrics.
{\rev However, it did not outperform comparative methods in terms of SEN adequately.} Finally, our proposed EAG-RS is observed to outperform all competing methods in terms of all metrics.

\begin{table}[t!]
\caption{Comparisons of ASD classification performances of the proposed method and state-of-the-art methods on the ABIDE dataset. Note that the entries corresponding to each method are based on the results reported in their respective manuscript.}
\label{table4}
\centering{
\setlength{\tabcolsep}{2pt}
\begin{tabular}{lcccccc}\\\toprule
Models & \multirow{1}{*}{AUC} & \multirow{1}{*}{ACC (\%)} & \multirow{1}{*}{SEN (\%)} & \multirow{1}{*}{SPEC (\%)}\\\toprule
% Nielsen et al.~\cite{ghiassian2016using}& - & $60.00$ & $62.00$ & $58.00$\\
% Ghiassian et al.~\cite{ghiassian2016using}& - & $65.00$ & $71.30$ & $58.30$\\
LSTM~\cite{dvornek2017identifying}& - & $68.50$ & - & -\\
DNN~\cite{heinsfeld2018identification}& - & $70.00$ & $\mathbf{74.00}$ & $63.00$\\
GCNs~\cite{parisot2018disease}& $0.750$ & $70.40$ & - & -\\
% Abraham et al.~\cite{abraham2017deriving}& - & $66.90$ & $78.30$ & $53.20$\\
% ASD-DiagNet~\cite{eslami2019asd}& - & $70.30$ & $68.30$ & $72.20$\\
% Almuqhim et al.~\cite{almuqhim2021asd}& - & $70.80$ & $62.00$ & $79.10$\\
% Sherkatghanad et al.~\cite{sherkatghanad2020automated}& $0.700$ & $70.22$ & $77.46$ & $73.55$\\
Extra-Trees+SVM~\cite{liu2020attentional}& - &$72.20$ & $68.80$ & $75.40$\\
MTFS-EM~\cite{liu2020improved} & $0.722$ & $69.39$ & $62.50$ & $74.06$\\
\midrule
\multirow{1}{*}{EAG-RS}& $\mathbf{0.760}$ & $\mathbf{73.71}$& ${64.56}$ & $\mathbf{80.74}$\\\bottomrule
\end{tabular}}
% \end{threeparttable}
\end{table}

\begin{table}[t!]
\caption{\rev Comparison of ASD identification performances of the proposed method and state-of-the-art methods on the ABIDE dataset. Note that the entries corresponding to each method are reimplemented using the same experimental configurations as those in our experiments.}
\label{table4_1}
\centering{
\setlength{\tabcolsep}{1pt}
\scalebox{0.9}{\begin{tabular}{lcccccc}\\\toprule
Models & \multirow{1}{*}{AUC} & \multirow{1}{*}{ACC (\%)} & \multirow{1}{*}{SEN (\%)} & \multirow{1}{*}{SPEC (\%)}\\\toprule
{\rev BrainNetCNN~\cite{kawahara2017brainnetcnn}}&{\rev 0.686$\pm$0.030}&{\rev 62.12$\pm$3.12}&{\rev 61.53$\pm$7.09}&{\rev 62.43$\pm$8.68}\\

{\rev BrainGNN~\cite{li2021braingnn}}&{\rev 0.627$\pm$0.040}&{\rev 61.13$\pm$3.54}&{\rev 58.60$\pm$9.25}&{\rev 63.37$\pm$8.21}\\
% Brain Network Transformer
{\rev BrainNetTF~\cite{kan2022brain}}&{\rev 0.700$\pm$0.026}&{\rev 65.60$\pm$3.34}&{\rev 64.38$\pm$3.26}&{\rev 66.42$\pm$4.69}\\
\midrule
\multirow{1}{*}{EAG-RS}& \textbf{0.760$\pm$0.033} & \textbf{73.71$\pm$2.83} & \textbf{64.56$\pm$8.36} & \textbf{80.74$\pm$8.28}\\\bottomrule
\end{tabular}}}
% \end{threeparttable}
\end{table}

\subsubsection{Comparison of feature selection performance}
Table \ref{table3} indicates that the subset-based method (\ie, LASSO) {\rev outperforms} the ranking-based method{\rev s} (\ie, $t$-test {\rev and RFE}). {\rev However, ASD-DiagNet performed better than conventional FS approaches, except in terms of SEN.} Although {\rev comparative} FS methods are used to select important features and remove redundant ones to improve classification performance, their performances are still lower than the proposed method without feature selection (EAG-RS w/o $\psi$). The proposed method with FS module (EAG-RS) outperformed all other methods {\rev in terms of} all metrics. 
Based on these promising results, we conclude that {\rev the steps adopted in the proposed EAG-RS} play pivotal roles in classifying TD and ASD. The two ROI-representative vectors can extract appropriate features. Moreover, they are based on the connection-wise relevance score, which not only {\rev enables diagnosis based on the original feature representations of FC but also serves as criteria for the extraction of diagnosis-informative ROIs based on feature selection.}

We further report and compare the classification results obtained from state-of-the-art methods on the ABIDE dataset to demonstrate the superiority of our proposed EAG-RS, as illustrated in Table~\ref{table4} {\rev and Table~\ref{table4_1}. In Table~\ref{table4_1}, a fair comparison is ensured by re-implementing all the methods using the same experimental configurations as those used in our study.}

\begin{figure}[t!]
\centering
\includegraphics[width=.45\textwidth]{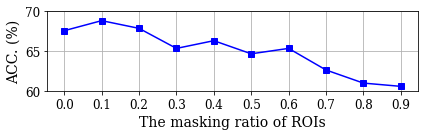}
\caption{\rev Effectiveness of ROI-masking ratio, $q$, of the proposed framework on the ABIDE dataset.}
\label{fig:training_curve}
\end{figure}

\section{Discussion}
\subsection{The ratio of random ROI-level masking}
{\rev First, the ratio of ROI-level masking is varied from 0 to 0.9 at intervals of 0.1. The corresponding performance results are presented in Fig.~4. When the ROI-level masking is $q=0.1$ and $q=0.2$, the performance is better than that obtained without ROI-level masking, indicating a positive influence of ROI-level masking on diagnostic performance. For this reason, $q=0.1$ is selected for subsequent experiments, as it corresponds to the highest performance quality.}

\subsection{Ablation Study}
{\rev We conduct additional experiments to validate the effectiveness of the proposed framework. We estimate features via six ablation cases in the context of a classification task. In Case I, the ROI selection network is removed and only FC features are used to classify the brain diseases using the multi-layer perceptron (MLP) (Case I = Baseline).} 
In Case II, the ROI selection network is removed and an estimated ROI representative vector, $\mathbf{f}_v$, is used to classify brain diseases. Under this setting, the number of dimensions {\rev is different} from {\rev that in} the original FC; thus, brain diseases are classified using an SVM. In Case III, the other estimated ROI representative vector, $\mathbf{f}_c$, is used {\rev for classification}. In Case IV, the original FC is used and the two ROI representative vectors (\ie, $\mathbf{f}=[\mathbf{f}_v||\mathbf{f}_c]$) are concatenated. Henceforth, we use an ROI selection network. Therefore, the original FC is implemented along with $\mathbf{f}_v$ (Case V), $\mathbf{f}_c$ (Case VI), and the concatenation of the two ROI representative vectors, $\mathbf{f}$ (EAG-RS).

As reported in Table~\ref{table2}, the proposed framework achieve the best diagnostic performance among all ablation cases. Cases without an ROI selection network (Case I, II, and III) are observed to exhibit lower classification performance. When independent representative vectors, $\mathbf{f}_v$ (Case II) and $\mathbf{f}_c$ (Case III), are used, slightly lower performance than that of the original FC (Case I) is observed. Therefore, the features are combined using concatenation to confirm the effectiveness of the representative features (Case IV), which improves the performance. However, {\rev the performance is observed to be degraded when each representative vector is used with an ROI selection network (Case V and VI).}
Thus, the count and value information aid the extraction of diagnosis-informative ROI information, which improves classification performance by removing redundant and irrelevant features of the original FC.

\begin{figure*}[t!]
\centering\vspace{-.3cm}
\includegraphics[width=\textwidth]{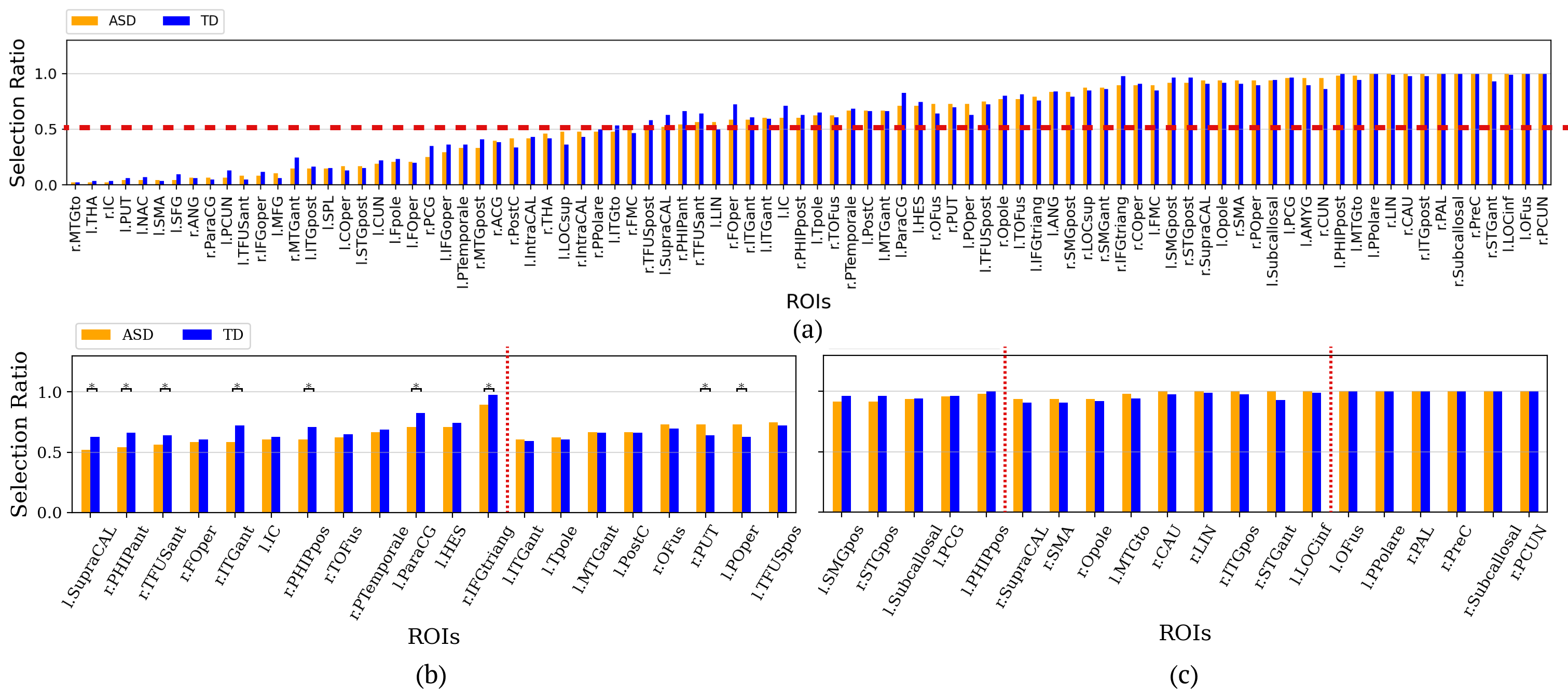}
\caption{(a) Visualization of the selection ratio (SR) of ROIs for each group at an ROI-level. Although we used 110 ROIs are used for this analysis, few had values of zero (\ie, not selected). In addition, we set $0.5$ as the threshold to consider the general patterns in SR. (b) The list of brain regions in each group (0.5 < SR < 0.75). The left and right sides of the red vertical line correspond to the TD and ASD groups, respectively. (c) The list of brain regions in each group (SR > 0.75). The left, middle, and right portions of the red vertical line correspond to the TD, ASD, and common groups, respectively.}
\label{fig4}
\end{figure*}

\begin{figure}[t!]
\centering
{\includegraphics[width=\columnwidth]{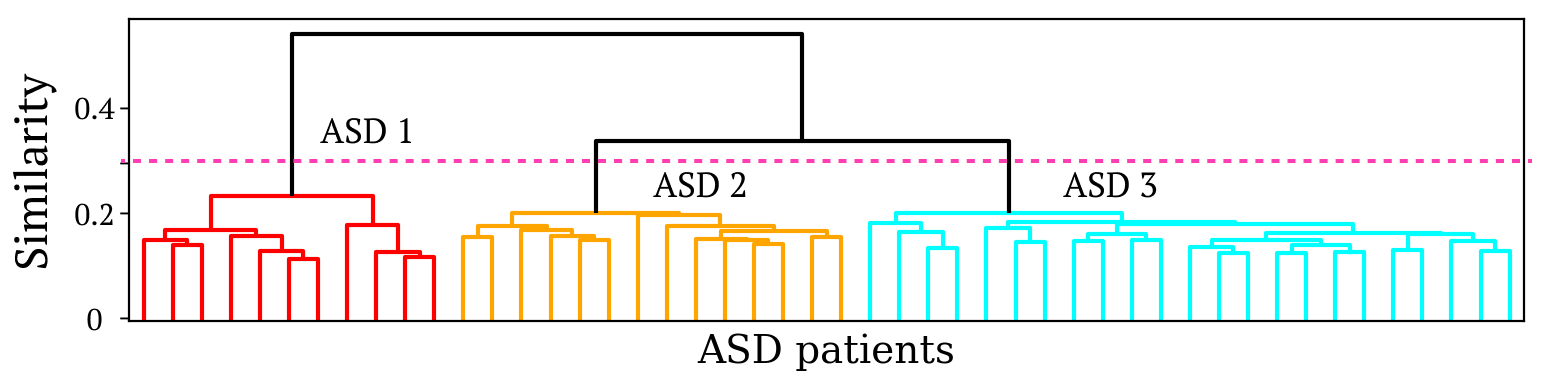}}
\caption{The Y-axis represents the similarity between patients, \ie, the shorter the distance, the greater the similarity between patients{\revv —the similarity threshold is set at 0.3}. The yellow, green, and red lines represent the clustered subtypes of ASD.}
\label{fig5}
\end{figure}

\begin{figure*}[t!]
\centering\vspace{-.3cm}
\includegraphics[width=\textwidth]{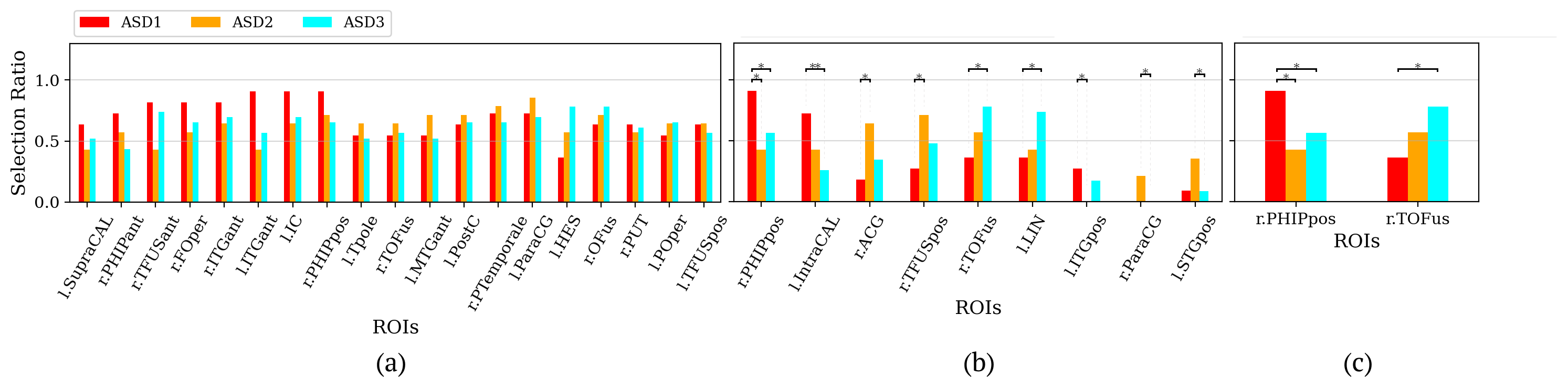}
\caption{(a) Visualization of the SR of ROIs corresponding to the three subtypes of ASD at ROI-level. The list of brain regions in the ASD group corresponds to $0.5<\text{SR}<0.75$. (b) Statistical test results for three subtypes of ASD analysis using ROIs selected by the ROI selection network. {\rev Only statistically significant ROIs are visualized.} Note that $*, **$ represent $p<0.05$ and $p<0.01$, respectively. (c) The list of common brain regions captured between (a) and (b).}
\label{fig6}
\end{figure*}

\begin{figure}[t!]
\centering\vspace{-.3cm}
\includegraphics[width=.5\textwidth]{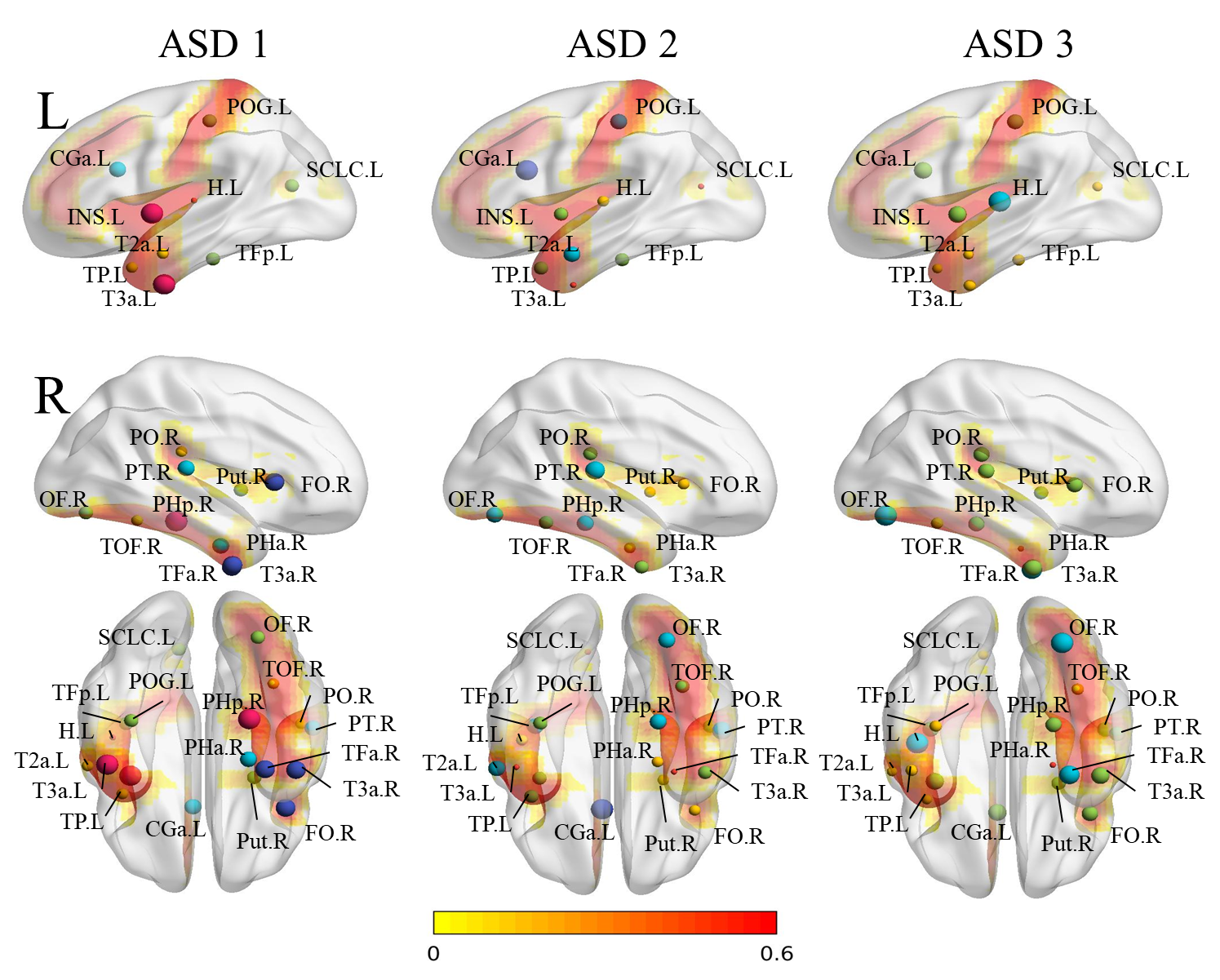}
\caption{\rev Visualization of 3D brain image representing SRs for each ASD subtype. Each circular marker indicates a specific brain region, and it is assigned a color based on the range of SR (SR < 0.5 (red), 0.5 < SR < 0.6 (yellow), 0.6 < SR < 0.7 (light green), 0.7 < SR < 0.8 (cyan), 0.8 < SR < 0.9 (dark blue), SR > 0.9 (magenta)).}
\label{fig7}
\end{figure}

\begin{table}[t!]
\caption{Averaged classification performances for ASD and TD in the ablation study.}
\label{table2}
\centering{
\setlength{\tabcolsep}{3pt}
\begin{tabular}{ccccccc}\\\toprule
    Case &\multirow{1}{*}{AUC} & \multirow{1}{*}{ACC (\%)} & \multirow{1}{*}{SEN (\%)} & \multirow{1}{*}{SPEC (\%)}\\\toprule
    I & {0.757$\pm$0.040} & {69.76$\pm$3.45} & {57.82$\pm$6.64} & {80.22$\pm$4.54}\\
    II & {0.552$\pm$0.020} & {48.53$\pm$4.58} & {55.35$\pm$2.19} & {55.75$\pm$2.41}\\
    III & {0.666$\pm$0.029} & {59.35$\pm$2.91} & {50.32$\pm$9.62} & {68.59$\pm$7.33}\\
    IV & \textbf{0.760$\pm$0.018} & {71.01$\pm$1.77} & \textbf{65.60$\pm$7.50} & {75.40$\pm$6.59}\\
    V &{0.747$\pm$0.028} & {70.64$\pm$3.28} & {65.25$\pm$5.04} & {75.29$\pm$5.29}\\
    VI &\textbf{0.760$\pm$0.025} & {71.09$\pm$3.29} & {64.46$\pm$5.74} & {78.26$\pm$6.01}\\\midrule
    EAG-RS &\textbf{0.760$\pm$0.033} & \textbf{73.71$\pm$2.83} & {64.56$\pm$8.36} & \textbf{80.74$\pm$8.28}\\
    \bottomrule
\end{tabular}}
\end{table}

\subsection{Analysis of ROI Selection Network}
{\rev The best performing model on the validation dataset is analyzed further. On average, 22 ROIs (median, 23) are selected for the ASD group, and 40 ROIs (median, 43) from the TD group.}
To visualize the selection ratio (SR) of each ROI, the ROIs selected from the $\psi$ module are enumerated and the total number is divided by the number of ROIs and the number of subjects in each group (Fig.~\ref{fig4}).
In particular, ROIs with SR exceeding 0.5 as considered (Fig.~\ref{fig4}(a)). These ROIs are further categorized into two cases: i) $0.5<\text{SR}<0.75$ (Fig.~\ref{fig4}(b)) and ii) $\text{SR}>0.75$ (Fig.~\ref{fig4}(c)).
The majority of selected brain regions are associated with ASD. As depicted in Fig.~\ref{fig4}(b), 12 brain regions lie on the left side of the red vertical line with slightly higher SR in the TD group, while eight regions lie on the right side of the vertical line with higher SR in the ASD group.
In Fig.~\ref{fig4}(c), five brain regions lie on the left side of the vertical line with slightly higher SR in the TD group, while nine regions lie on the middle side of the vertical line with higher SR in the ASD group.
The SR patterns reveal that 12 specific brain regions are always selected, with six brain regions on the right side of the vertical line (Fig.~\ref{fig4}(c)) chosen consistently in both the TD and ASD groups. In the TD group, the ``l.PHIPpos" region is selected 100\% of the time, while in the ASD group, five specific brain regions (``r.CAU," ``r.LIN," ``l.ITGpos," ``r.STGant," and ``r.LOCinf") are chosen consistently.

Group analysis is performed based on these selected ROIs to perform a neuroscientific analysis of ASD and TD groups at the ROI level. Subsequently, we identify nine brain regions with significantly different selection frequency between the two groups by measuring the difference (<0.05). {\rev Remarkably, the proposed framework, trained without prior knowledge, is observed to identify brain regions highly related to existing neuroscience studies. Specifically, it identifies the following brain regions: 'l.SupraCAL,' 'r.PHIPant,' 'r.TFUSant,' 'r.ITGant,' 'r.PHIPpos,' 'l.ParaCG,' 'r.IFGtriang,' 'r.PUT,' 'l.POper.'
These regions are marked with asterisks (*) in Fig.~\ref{fig4}(b). Notably, the left supracalcarine cortex (l.SupraCAL), situated in the visual cortex, is involved in various visual processes such as discerning object shape, size, and color, and motion perception~\cite{dichter2009autism}. Similarly, the right inferior temporal gyrus (r.ITGant) plays a similar role~\cite{kim2021overconnectivity}. In addition, the right temporal fusiform cortex (r.TFUSant) in the temporal lobe is responsible for facial processing and recognition. It distinguishes facial features and supports social interactions and recognition skills~\cite{van2008neurons}. The right parahippocampal gyrus (r.PHIPant and r.PHIPpos) in the medial temporal lobe is vital to memory, spatial navigation, and emotional processing~\cite{monk2009abnormalities}. The left paracingulate gyrus (l.ParaCG), which is a part of the cingulate cortex, contributes to various cognitive and emotional functions. The right putamen (r.PUT), a basal ganglia structure, participates in motor control, procedural learning, reward processing, attention, and cognition. The left parietal operculum cortex (l.POper) has broad involvement in somatosensory processing, language, and multisensory integration, supporting sensory perception, communication, and body awareness. Finally, the right inferior frontal gyrus triangular (r.IFGtriang) region contains Broca's area, essential for language processing like production, comprehension, and sophisticated inhibitory control. This region's crucial role in language-related functions and cognitive processes is well-documented~\cite{yuk2018you}. This confirmation of the biological relevance and interpretability of our findings further supports the conclusion that it is effective in identifying important brain regions associated with target disorders.}

\begin{table}[t!]
\centering{
\caption{\rev Characteristics of demographic, psychological, and educational assessments for TD and ASD subtypes.}
\label{table5}
% \resizebox{\columnwidth}{!}{
\setlength{\tabcolsep}{2pt}
\begin{tabular}{lcccccc}
    \toprule
    Categories & TD & ASD1 & ASD2 & ASD3 & F & $p$-value (\textgreater{}F) \\\midrule
    Age (years) & $15.46$ & $12.90$ & $15.69$ & $15.24$ & $0.75$ & $0.528$ \\\midrule
    Male (\%) & $76.92$ & $81.82$ & $92.86$ & $87.50$ & 
    \multirow{2}{*}{$0.80$} & \multirow{2}{*}{$0.497$} \\
    Female (\%) & $23.08$ & $18.18$ & $7.14$ & $12.50$ &  &  \\\midrule
    % Handedness &  &  &  &  &  &  \\
    % Left-handed (\%) & $7.69$ & $9.09$ & $0.00$ & $18.75$ & \multirow{3}{*}{$0.34$} & \multirow{3}{*}{$0.797$} \\
    % Right-handed (\%) & $80.00$ & $72.73$ & $100.00$ & $68.75$ &  &  \\
    % Ambidextrous (\%) & $12.31$ & $18.18$ & $0.00$ & $12.50$ &  &  \\\midrule
    FIQ & $113.18$ & $97.95$ & $99.61$ & $94.81$ & $10.31$ & \textless $0.001$ \\
    VIQ & $114.14$ & $101.09$ & $99.86$ & $96.69$ & $5.89$ & \textless $0.001$ \\
    PIQ & $109.63$ & $98.64$ & $100.36$ & $92.75$ & $9.42$ & \textless $0.001$ \\ \bottomrule
\end{tabular}%
}
\begin{tablenotes}
    \tiny \item ASD, autism spectrum disorder; F, $F$-value; FIQ, full-scale intelligence quotient;\\
    VIQ, verbal IQ; PIQ, performance IQ;
\end{tablenotes}
%		\end{threeparttable}
\end{table}

\subsection{Clustering Subtypes in Autism Spectrum Disorder}

{\rev ASD is known for its diverse and heterogeneous nature, with various characteristics associated with ASD-related brain regions, and varying {\rev symptom severity levels and comorbidities}~\cite{moradi2017predicting}. As a result, several previous research projects in ASD have aimed to identify distinct behavioral subtypes within ASD populations.
In this study, we consider such heterogeneity among ASD groups during ROI selection. 
To explore this heterogeneity further, subtype analysis is performed using hierarchical clustering with ward linkage, a commonly used method in neuroimaging. The results reveal three subtypes of ASD (Fig.~\ref{fig5}), {\rev each with unique characteristics and functional connectivity patterns}.
Further, we investigate demographic information, such as age and gender, of each identified ASD subtype using statistical tests (Table~\ref{table5}). In Table~\ref{table5}, the average age of TD individuals is observed to be 15.5 years. Among the clustered ASD subtypes, ASD1 exhibits an average age of 12.9 years; ASD2, 15.7 years; and ASD3, 15.2 years. In the case of gender distribution, the TD group exhibits a female proportion of 23.1\%. For the identified ASD subtypes, ASD1 exhibits a female proportion of 18.2\%; ASD2, 7.1\%; and ASD3, 12.5\%.
Further, in terms of Full-Scale Intelligence Quotient (FIQ), Verbal Intelligence Quotient (VIQ), and Performance Intelligence Quotient (PIQ), TD individuals exhibit average scores of 113.2, 114.1, and 109.6, respectively. 
In comparison, the ASD1 subtype exhibits average scores of 98.0 (FIQ), 101.1 (VIQ), and 98.6 (PIQ), respectively; ASD2, 99.6 (FIQ), 99.9 (VIQ), and 100.4 (PIQ), respectively; and ASD3, 94.8 (FIQ), 96.7 (VIQ), and 92.8 (PIQ), respectively.
This analysis provides deep insight into the potential relationships between the identified subtypes and various demographic factors, as well as their associations with psychological and educational assessments.}

{\rev{\rev Based on our analysis of the selected ROIs depicted in Fig.~\ref{fig4}(b), they are sorted based on their SRs for each subtype of ASD.} In addition, the significant regions {\rev associated with} ASD subtypes are visualized in Fig.~\ref{fig6}(b).
Two common regions are observed---the right posterior parahippocampus (r.PHIPpos) and right temporal occipital fusiform cortex (r.TOFus). These are {\rev known to be} associated with ASD, as depicted in Fig.~\ref{fig6}(a) and Fig~\ref{fig6}(b). {\rev Interestingly, as presented in Fig.~\ref{fig6}, we observed} different SRs for different ASD subtypes in {\rev these} regions. {\rev Finally, to provide a more intuitive understanding, the SRs corresponding to each ASD subtype are mapped onto 3D brain images and visualized in Fig.~\ref{fig7}. In Fig.~\ref{fig7}, the brain regions are represented by circular markers, and the color of each marker corresponds to the range of SR values for that specific region. These results provide beneficial insights into ASD subtyping from a neuroscientific viewpoint. The proposed ROI selection network successfully identifies and differentiates specific brain regions associated with different ASD subtypes. We believe that this capability enhances our understanding of the biological basis of the heterogeneity within the ASD population and may have significant implications for the advancement of subtype analysis in autism research.}}

\section{Conclusion}
{\rev {In this work, we propose a novel explainability-guided ROI selection (EAG-RS) framework that dynamically selects informative features at the ROI-level for brain disease diagnosis.
Our EAG-RS framework learns inter-regional relationships using random seed-based network masking to estimate non-linear relationships, representing other neighboring connections to restore masked seed-ROI connections. We also estimated connection-wise relevance scores to explore high-order relationships between FCs using LRP. Finally, we utilized the estimated non-linear high-order FCs to select diagnosis-informative ROIs and diagnose brain disease simultaneously. To demonstrate its validity, ASD diagnosis was performed using the proposed EAG-RS framework on the ABIDE dataset.
Furthermore, the cluster subtypes for ASD were identified based on individually selected ROIs. The results demonstrate that our EAG-RS framework provides new neuroscientific insights into ASD subtypes and their biomarkers.}

{\rev However, this study suffers from the following practical limitations.
First, regarding the architectural design, MLPs were used for the encoder-decoder structure in inter-regional relation learning and the ROI selection network. Although MLPs provide flexibility and expressiveness, they involve a high number of tunable parameters. Given the scarcity of neuroimaging data and labels, optimization of such MLP-based architectures may be challenging, requiring strong regularization. To address this limitation, alternative approaches may be explored, \eg, incorporating convolutional neural networks and transformers into each module of the proposed framework. Second, as the sole focus of this study was the use of FC for ASD diagnosis, it did not incorporate other neuroimaging modalities or clinical information. Integrating multiple modalities and clinical data could potentially provide complementary insights and improve the accuracy and robustness of ASD diagnosis. Addressing the abovementioned limitations in the future will contribute to a more comprehensive understanding of the application of FC in ASD diagnosis and enhance the effectiveness and interpretability of the proposed framework.}

\bibliographystyle{IEEEtran}
\bibliography{main}}

\begin{thebibliography}{10}
\providecommand{\url}[1]{#1}
\csname url@samestyle\endcsname
\providecommand{\newblock}{\relax}
\providecommand{\bibinfo}[2]{#2}
\providecommand{\BIBentrySTDinterwordspacing}{\spaceskip=0pt\relax}
\providecommand{\BIBentryALTinterwordstretchfactor}{4}
\providecommand{\BIBentryALTinterwordspacing}{\spaceskip=\fontdimen2\font plus
\BIBentryALTinterwordstretchfactor\fontdimen3\font minus
  \fontdimen4\font\relax}
\providecommand{\BIBforeignlanguage}[2]{{%
\expandafter\ifx\csname l@#1\endcsname\relax
\typeout{** WARNING: IEEEtran.bst: No hyphenation pattern has been}%
\typeout{** loaded for the language `#1'. Using the pattern for}%
\typeout{** the default language instead.}%
\else
\language=\csname l@#1\endcsname
\fi
#2}}
\providecommand{\BIBdecl}{\relax}
\BIBdecl

\bibitem{ecker2010describing}
C.~Ecker, A.~Marquand, J.~Mour{\~a}o-Miranda, P.~Johnston, E.~M. Daly, M.~J.
  Brammer, S.~Maltezos, C.~M. Murphy, D.~Robertson, S.~C. Williams
  \emph{et~al.}, ``Describing the brain in autism in five dimensions—magnetic
  resonance imaging-assisted diagnosis of autism spectrum disorder using a
  multiparameter classification approach,'' \emph{Journal of Neuroscience},
  vol.~30, no.~32, pp. 10\,612--10\,623, 2010.

\bibitem{buescher2014costs}
A.~V. Buescher, Z.~Cidav, M.~Knapp, and D.~S. Mandell, ``Costs of autism
  spectrum disorders in the united kingdom and the united states,'' \emph{JAMA
  pediatrics}, vol. 168, no.~8, pp. 721--728, 2014.

\bibitem{fernell2013early}
E.~Fernell, M.~A. Eriksson, and C.~Gillberg, ``Early diagnosis of autism and
  impact on prognosis: a narrative review,'' \emph{Clinical epidemiology},
  vol.~5, p.~33, 2013.

\bibitem{zwaigenbaum2015early}
L.~Zwaigenbaum, M.~L. Bauman, W.~L. Stone, N.~Yirmiya, A.~Estes, R.~L. Hansen,
  J.~C. McPartland, M.~R. Natowicz, R.~Choueiri, D.~Fein \emph{et~al.}, ``Early
  identification of autism spectrum disorder: Recommendations for practice and
  research,'' \emph{Pediatrics}, vol. 136, no. Supplement\_1, pp. S10--S40,
  2015.

\bibitem{abraham2017deriving}
A.~Abraham, M.~P. Milham, A.~Di~Martino, R.~C. Craddock, D.~Samaras,
  B.~Thirion, and G.~Varoquaux, ``Deriving reproducible biomarkers from
  multi-site resting-state data: An autism-based example,'' \emph{NeuroImage},
  vol. 147, pp. 736--745, 2017.

\bibitem{kim2016deep}
J.~Kim, V.~D. Calhoun, E.~Shim, and J.-H. Lee, ``Deep neural network with
  weight sparsity control and pre-training extracts hierarchical features and
  enhances classification performance: Evidence from whole-brain resting-state
  functional connectivity patterns of schizophrenia,'' \emph{Neuroimage}, vol.
  124, pp. 127--146, 2016.

\bibitem{brier2012loss}
M.~R. Brier, J.~B. Thomas, A.~Z. Snyder, T.~L. Benzinger, D.~Zhang, M.~E.
  Raichle, D.~M. Holtzman, J.~C. Morris, and B.~M. Ances, ``Loss of
  intranetwork and internetwork resting state functional connections with
  alzheimer's disease progression,'' \emph{Journal of Neuroscience}, vol.~32,
  no.~26, pp. 8890--8899, 2012.

\bibitem{cribben2012dynamic}
I.~Cribben, R.~Haraldsdottir, L.~Y. Atlas, T.~D. Wager, and M.~A. Lindquist,
  ``Dynamic connectivity regression: determining state-related changes in brain
  connectivity,'' \emph{Neuroimage}, vol.~61, no.~4, pp. 907--920, 2012.

\bibitem{dvornek2017identifying}
N.~C. Dvornek, P.~Ventola, K.~A. Pelphrey, and J.~S. Duncan, ``Identifying
  autism from resting-state fmri using long short-term memory networks,'' in
  \emph{International Workshop on Machine Learning in Medical Imaging}.\hskip
  1em plus 0.5em minus 0.4em\relax Springer, 2017, pp. 362--370.

\bibitem{kang2018probabilistic}
E.~Kang and H.-I. Suk, ``Probabilistic source separation on resting-state fmri
  and its use for early mci identification,'' in \emph{International Conference
  on Medical Image Computing and Computer-Assisted Intervention}.\hskip 1em
  plus 0.5em minus 0.4em\relax Springer, 2018, pp. 275--283.

\bibitem{jeon2020enriched}
E.~Jeon, E.~Kang, J.~Lee, J.~Lee, T.-E. Kam, and H.-I. Suk, ``Enriched
  representation learning in resting-state fmri for early mci diagnosis,'' in
  \emph{International Conference on Medical Image Computing and
  Computer-Assisted Intervention}.\hskip 1em plus 0.5em minus 0.4em\relax
  Springer, 2020, pp. 397--406.

\bibitem{chen2016high}
X.~Chen, H.~Zhang, Y.~Gao, C.-Y. Wee, G.~Li, D.~Shen, and A.~D.~N. Initiative,
  ``High-order resting-state functional connectivity network for mci
  classification,'' \emph{Human brain mapping}, vol.~37, no.~9, pp. 3282--3296,
  2016.

\bibitem{zhao2020diagnosis}
F.~Zhao, Z.~Chen, I.~Rekik, S.-W. Lee, and D.~Shen, ``Diagnosis of autism
  spectrum disorder using central-moment features from low-and high-order
  dynamic resting-state functional connectivity networks,'' \emph{Frontiers in
  neuroscience}, vol.~14, 2020.

\bibitem{biswal2012resting}
B.~B. Biswal, ``Resting state fmri: a personal history,'' \emph{Neuroimage},
  vol.~62, no.~2, pp. 938--944, 2012.

\bibitem{lee2003report}
L.~Lee, L.~M. Harrison, and A.~Mechelli, ``A report of the functional
  connectivity workshop, dusseldorf 2002,'' \emph{Neuroimage}, vol.~19, no.~2,
  pp. 457--465, 2003.

\bibitem{van2010exploring}
M.~P. Van Den~Heuvel and H.~E.~H. Pol, ``Exploring the brain network: a review
  on resting-state fmri functional connectivity,'' \emph{European
  neuropsychopharmacology}, vol.~20, no.~8, pp. 519--534, 2010.

\bibitem{wager2017imaging}
T.~D. Wager and C.-W. Woo, ``Imaging biomarkers and biotypes for depression,''
  \emph{Nature medicine}, vol.~23, no.~1, pp. 16--17, 2017.

\bibitem{parisot2018disease}
S.~Parisot, S.~I. Ktena, E.~Ferrante, M.~Lee, R.~Guerrero, B.~Glocker, and
  D.~Rueckert, ``Disease prediction using graph convolutional networks:
  application to autism spectrum disorder and alzheimer’s disease,''
  \emph{Medical image analysis}, vol.~48, pp. 117--130, 2018.

\bibitem{heinsfeld2018identification}
A.~S. Heinsfeld, A.~R. Franco, R.~C. Craddock, A.~Buchweitz, and F.~Meneguzzi,
  ``Identification of autism spectrum disorder using deep learning and the
  abide dataset,'' \emph{NeuroImage: Clinical}, vol.~17, pp. 16--23, 2018.

\bibitem{suk2016state}
H.-I. Suk, C.-Y. Wee, S.-W. Lee, and D.~Shen, ``State-space model with deep
  learning for functional dynamics estimation in resting-state fmri,''
  \emph{NeuroImage}, vol. 129, pp. 292--307, 2016.

\bibitem{guo2017diagnosing}
X.~Guo, K.~C. Dominick, A.~A. Minai, H.~Li, C.~A. Erickson, and L.~J. Lu,
  ``Diagnosing autism spectrum disorder from brain resting-state functional
  connectivity patterns using a deep neural network with a novel feature
  selection method,'' \emph{Frontiers in neuroscience}, vol.~11, p. 460, 2017.

\bibitem{eslami2019asd}
T.~Eslami, V.~Mirjalili, A.~Fong, A.~R. Laird, and F.~Saeed, ``Asd-diagnet: a
  hybrid learning approach for detection of autism spectrum disorder using fmri
  data,'' \emph{Frontiers in neuroinformatics}, vol.~13, p.~70, 2019.

\bibitem{wang2019identification}
C.~Wang, Z.~Xiao, B.~Wang, and J.~Wu, ``Identification of autism based on
  svm-rfe and stacked sparse auto-encoder,'' \emph{Ieee Access}, vol.~7, pp.
  118\,030--118\,036, 2019.

\bibitem{rakic2020improving}
M.~Raki{\'c}, M.~Cabezas, K.~Kushibar, A.~Oliver, and X.~Llado, ``Improving the
  detection of autism spectrum disorder by combining structural and functional
  mri information,'' \emph{NeuroImage: Clinical}, vol.~25, p. 102181, 2020.

\bibitem{jung2021inter}
W.~Jung, D.-W. Heo, E.~Jeon, J.~Lee, and H.-I. Suk, ``Inter-regional high-level
  relation learning from functional connectivity via self-supervision,'' in
  \emph{International Conference on Medical Image Computing and
  Computer-Assisted Intervention}.\hskip 1em plus 0.5em minus 0.4em\relax
  Springer, 2021, pp. 284--293.

\bibitem{wee2012identification}
C.-Y. Wee, P.-T. Yap, D.~Zhang, K.~Denny, J.~N. Browndyke, G.~G. Potter, K.~A.
  Welsh-Bohmer, L.~Wang, and D.~Shen, ``Identification of mci individuals using
  structural and functional connectivity networks,'' \emph{Neuroimage},
  vol.~59, no.~3, pp. 2045--2056, 2012.

\bibitem{tibshirani1996regression}
R.~Tibshirani, ``Regression shrinkage and selection via the lasso,''
  \emph{Journal of the Royal Statistical Society: Series B (Methodological)},
  vol.~58, no.~1, pp. 267--288, 1996.

\bibitem{wang2019graph}
M.~Wang, B.~Jie, W.~Bian, X.~Ding, W.~Zhou, Z.~Wang, and M.~Liu, ``Graph-kernel
  based structured feature selection for brain disease classification using
  functional connectivity networks,'' \emph{IEEE Access}, vol.~7, pp.
  35\,001--35\,011, 2019.

\bibitem{zhu2019hybrid}
Q.~Zhu, H.~Li, J.~Huang, X.~Xu, D.~Guan, and D.~Zhang, ``Hybrid functional
  brain network with first-order and second-order information for
  computer-aided diagnosis of schizophrenia,'' \emph{Frontiers in
  neuroscience}, vol.~13, p. 603, 2019.

\bibitem{zhang2016topographical}
H.~Zhang, X.~Chen, F.~Shi, G.~Li, M.~Kim, P.~Giannakopoulos, S.~Haller, and
  D.~Shen, ``Topographical information-based high-order functional connectivity
  and its application in abnormality detection for mild cognitive impairment,''
  \emph{Journal of Alzheimer's Disease}, vol.~54, no.~3, pp. 1095--1112, 2016.

\bibitem{zhao2021constructing}
F.~Zhao, X.~Zhang, K.-H. Thung, N.~Mao, S.-W. Lee, and D.~Shen, ``Constructing
  multi-view high-order functional connectivity networks for diagnosis of
  autism spectrum disorder,'' \emph{IEEE Transactions on Biomedical
  Engineering}, vol.~69, no.~3, pp. 1237--1250, 2021.

\bibitem{zhang2017hybrid}
Y.~Zhang, H.~Zhang, X.~Chen, S.-W. Lee, and D.~Shen, ``Hybrid high-order
  functional connectivity networks using resting-state functional mri for mild
  cognitive impairment diagnosis,'' \emph{Scientific reports}, vol.~7, no.~1,
  p. 6530, 2017.

\bibitem{bach2015pixel}
S.~Bach, A.~Binder, G.~Montavon, F.~Klauschen, K.-R. M{\"u}ller, and W.~Samek,
  ``On pixel-wise explanations for non-linear classifier decisions by
  layer-wise relevance propagation,'' \emph{PloS one}, vol.~10, no.~7, p.
  e0130140, 2015.

\bibitem{di2014autism}
A.~Di~Martino, C.-G. Yan, Q.~Li, E.~Denio, F.~X. Castellanos, K.~Alaerts, J.~S.
  Anderson, M.~Assaf, S.~Y. Bookheimer, M.~Dapretto \emph{et~al.}, ``{The
  Autism Brain Imaging Data Exchange: Towards a Large-scale Evaluation of The
  Intrinsic Brain Architecture in Autism},'' \emph{{Molecular Psychiatry}},
  vol.~19, no.~6, pp. 659--667, 2014.

\bibitem{liu2020improved}
J.~Liu, Y.~Sheng, W.~Lan, R.~Guo, Y.~Wang, and J.~Wang, ``Improved asd
  classification using dynamic functional connectivity and multi-task feature
  selection,'' \emph{Pattern Recognition Letters}, vol. 138, pp. 82--87, 2020.

\bibitem{yan2017discriminating}
W.~Yan, S.~Plis, V.~D. Calhoun, S.~Liu, R.~Jiang, T.-Z. Jiang, and J.~Sui,
  ``Discriminating schizophrenia from normal controls using resting state
  functional network connectivity: A deep neural network and layer-wise
  relevance propagation method,'' in \emph{2017 IEEE 27th international
  workshop on machine learning for signal processing (MLSP)}.\hskip 1em plus
  0.5em minus 0.4em\relax IEEE, 2017, pp. 1--6.

\bibitem{azevedo2022deep}
T.~Azevedo, A.~Campbell, R.~Romero-Garcia, L.~Passamonti, R.~A. Bethlehem,
  P.~Li{\`o}, and N.~Toschi, ``A deep graph neural network architecture for
  modelling spatio-temporal dynamics in resting-state functional mri data,''
  \emph{Medical Image Analysis}, vol.~79, p. 102471, 2022.

\bibitem{zhao2022attention}
M.~Zhao, W.~Yan, N.~Luo, D.~Zhi, Z.~Fu, Y.~Du, S.~Yu, T.~Jiang, V.~D. Calhoun,
  and J.~Sui, ``An attention-based hybrid deep learning framework integrating
  brain connectivity and activity of resting-state functional mri data,''
  \emph{Medical image analysis}, vol.~78, p. 102413, 2022.

\bibitem{lin2022sspnet}
Q.-H. Lin, Y.-W. Niu, J.~Sui, W.-D. Zhao, C.~Zhuo, and V.~D. Calhoun, ``Sspnet:
  An interpretable 3d-cnn for classification of schizophrenia using phase maps
  of resting-state complex-valued fmri data,'' \emph{Medical Image Analysis},
  vol.~79, p. 102430, 2022.

\bibitem{dang2019novel}
S.~Dang and S.~Chaudhury, ``Novel relative relevance score for estimating brain
  connectivity from fmri data using an explainable neural network approach,''
  \emph{Journal of Neuroscience Methods}, vol. 326, p. 108371, 2019.

\bibitem{he2022masked}
K.~He, X.~Chen, S.~Xie, Y.~Li, P.~Doll{\'a}r, and R.~Girshick, ``Masked
  autoencoders are scalable vision learners,'' in \emph{Proceedings of the
  IEEE/CVF Conference on Computer Vision and Pattern Recognition}, 2022, pp.
  16\,000--16\,009.

\bibitem{gholizadeh2021model}
S.~Gholizadeh and N.~Zhou, ``Model explainability in deep learning based
  natural language processing,'' \emph{arXiv preprint arXiv:2106.07410}, 2021.

\bibitem{kingma2014adam}
D.~P. Kingma and J.~Ba, ``Adam: A method for stochastic optimization,''
  \emph{arXiv preprint arXiv:1412.6980}, 2014.

\bibitem{vincent2008extracting}
P.~Vincent, H.~Larochelle, Y.~Bengio, and P.-A. Manzagol, ``Extracting and
  composing robust features with denoising autoencoders,'' in \emph{Proceedings
  of the 25th international conference on Machine learning}, 2008, pp.
  1096--1103.

\bibitem{vincent2010stacked}
P.~Vincent, H.~Larochelle, I.~Lajoie, Y.~Bengio, P.-A. Manzagol, and L.~Bottou,
  ``Stacked denoising autoencoders: Learning useful representations in a deep
  network with a local denoising criterion.'' \emph{Journal of machine learning
  research}, vol.~11, no.~12, 2010.

\bibitem{pathak2016context}
D.~Pathak, P.~Krahenbuhl, J.~Donahue, T.~Darrell, and A.~A. Efros, ``Context
  encoders: Feature learning by inpainting,'' in \emph{Proceedings of the IEEE
  conference on computer vision and pattern recognition}, 2016, pp. 2536--2544.

\bibitem{mcnemar1947note}
Q.~McNemar, ``Note on the {S}ampling {E}rror of the {D}ifference between
  {C}orrelated {P}roportions or {P}ercentages,'' \emph{{Psychometrika}},
  vol.~12, no.~2, pp. 153--157, 1947.

\bibitem{guyon2002gene}
I.~Guyon, J.~Weston, S.~Barnhill, and V.~Vapnik, ``Gene selection for cancer
  classification using support vector machines,'' \emph{{Machine Learning}},
  vol.~46, pp. 389--422, 2002.

\bibitem{kawahara2017brainnetcnn}
J.~Kawahara, C.~J. Brown, S.~P. Miller, B.~G. Booth, V.~Chau, R.~E. Grunau,
  J.~G. Zwicker, and G.~Hamarneh, ``Brainnetcnn: Convolutional neural networks
  for brain networks; towards predicting neurodevelopment,'' \emph{NeuroImage},
  vol. 146, pp. 1038--1049, 2017.

\bibitem{li2021braingnn}
X.~Li, Y.~Zhou, N.~Dvornek, M.~Zhang, S.~Gao, J.~Zhuang, D.~Scheinost, L.~H.
  Staib, P.~Ventola, and J.~S. Duncan, ``Braingnn: Interpretable brain graph
  neural network for fmri analysis,'' \emph{Medical Image Analysis}, vol.~74,
  p. 102233, 2021.

\bibitem{kan2022brain}
X.~Kan, W.~Dai, H.~Cui, Z.~Zhang, Y.~Guo, and C.~Yang, ``Brain network
  transformer,'' \emph{Advances in Neural Information Processing Systems},
  vol.~35, pp. 25\,586--25\,599, 2022.

\bibitem{bengio2006greedy}
Y.~Bengio, P.~Lamblin, D.~Popovici, and H.~Larochelle, ``Greedy layer-wise
  training of deep networks,'' \emph{Advances in neural information processing
  systems}, vol.~19, 2006.

\bibitem{liu2020attentional}
Y.~Liu, L.~Xu, J.~Li, J.~Yu, and X.~Yu, ``Attentional connectivity-based
  prediction of autism using heterogeneous rs-fmri data from cc200 atlas,''
  \emph{Experimental neurobiology}, vol.~29, no.~1, p.~27, 2020.

\bibitem{dichter2009autism}
G.~S. Dichter, J.~N. Felder, and J.~W. Bodfish, ``Autism is characterized by
  dorsal anterior cingulate hyperactivation during social target detection,''
  \emph{Social cognitive and affective neuroscience}, vol.~4, no.~3, pp.
  215--226, 2009.

\bibitem{kim2021overconnectivity}
D.~Kim, J.~Y. Lee, B.~C. Jeong, J.-H. Ahn, J.~I. Kim, E.~S. Lee, H.~Kim, H.~J.
  Lee, and C.~E. Han, ``Overconnectivity of the right heschl's and inferior
  temporal gyrus correlates with symptom severity in preschoolers with autism
  spectrum disorder,'' \emph{Autism Research}, vol.~14, no.~11, pp. 2314--2329,
  2021.

\bibitem{van2008neurons}
I.~A. van Kooten, S.~J. Palmen, P.~von Cappeln, H.~W. Steinbusch, H.~Korr,
  H.~Heinsen, P.~R. Hof, H.~van Engeland, and C.~Schmitz, ``Neurons in the
  fusiform gyrus are fewer and smaller in autism,'' \emph{Brain}, vol. 131,
  no.~4, pp. 987--999, 2008.

\bibitem{monk2009abnormalities}
C.~S. Monk, S.~J. Peltier, J.~L. Wiggins, S.-J. Weng, M.~Carrasco, S.~Risi, and
  C.~Lord, ``Abnormalities of intrinsic functional connectivity in autism
  spectrum disorders,'' \emph{Neuroimage}, vol.~47, no.~2, pp. 764--772, 2009.

\bibitem{yuk2018you}
V.~Yuk, C.~Urbain, E.~W. Pang, E.~Anagnostou, D.~Buchsbaum, and M.~J. Taylor,
  ``Do you know what i’m thinking? temporal and spatial brain activity during
  a theory-of-mind task in children with autism,'' \emph{Developmental
  cognitive neuroscience}, vol.~34, pp. 139--147, 2018.

\bibitem{moradi2017predicting}
E.~Moradi, B.~Khundrakpam, J.~D. Lewis, A.~C. Evans, and J.~Tohka, ``Predicting
  symptom severity in autism spectrum disorder based on cortical thickness
  measures in agglomerative data,'' \emph{Neuroimage}, vol. 144, pp. 128--141,
  2017.

\end{thebibliography}

\end{document}